\documentclass[10pt,journal,cspaper,compsoc]{IEEEtran}

\ifCLASSOPTIONcompsoc
  \usepackage[nocompress]{cite}
\else
  \usepackage{cite}
\fi

\usepackage[dvips]{graphicx}
\usepackage{array}
\usepackage{threeparttable}
\usepackage{amsmath}
\usepackage{multirow}
\usepackage{booktabs}
\usepackage{graphicx}
\usepackage{tabularx}
\usepackage{float}
\usepackage{amsopn}
\usepackage{xcolor}
\usepackage{makecell}
\usepackage{caption}
\usepackage[ruled]{algorithm2e}
\usepackage{stfloats} 

\usepackage{subcaption,graphicx}

\makeatletter
\newcommand{\removelatexerror}{\let\@latex@error\@gobble}
\makeatother

\newcounter{appcount}
\newcommand{\appendicesname}
            {Appendix\ \thechapter  \Alph{appcount}}
\newcommand{\bookappendicesname}
            {Appendix\ \Alph{appcount}}
\newcommand{\chapterappendix}[1]
          {\par\setcounter{section}{0}
           \setcounter{equation}{0}
           \setcounter{table}{0}
           \setcounter{figure}{0}
          \addtocounter{appcount}{1}   \renewcommand{\theequation}{\thechapter\Alph{appcount}.\arabic{equation}}
          \renewcommand{\thetable}{\thechapter\Alph{appcount}.\arabic{table}}
          \renewcommand{\thefigure}{\thechapter\Alph{appcount}.\arabic{figure}}
           \setcounter{section}{\arabic{chapter}\Alph{section}}
           \if@openright\cleardoublepage\else\clearpage\fi
           \chapter*{\huge{\appendicesname}\newline\newline \Huge{#1}}
           \addcontentsline{toc}{section}{\thechapter\Alph{appcount} #1}
           \markright{\MakeUppercase{\appendicesname.\ { #1}}}}
\newcommand{\bookappendix}[1]
          {\par\setcounter{section}{0}
           \setcounter{equation}{0}
           \setcounter{table}{0}
           \setcounter{figure}{0}
          \addtocounter{appcount}{1}   \renewcommand{\theequation}{\Alph{appcount}.\arabic{equation}}
           \renewcommand{\thetable}{\Alph{appcount}.\arabic{table}}
             \renewcommand{\thefigure}{\Alph{appcount}.\arabic{figure}}
           \setcounter{section}{\arabic{chapter}\Alph{section}}
           \if@openright\cleardoublepage\else\clearpage\fi
           \chapter*{\huge{\bookappendicesname}\newline\newline \Huge{#1}}
           \addcontentsline{toc}{chapter}{\bookappendicesname #1}
          \markright{\MakeUppercase{\bookappendicesname.\ { #1}}} }
\newcounter{example}
\newcounter{property}
\newcommand{\ben}{\begin{equation}}
\newcommand{\een}{\end{equation}}
\newcommand{\bea}{\begin{eqnarray*}}
\newcommand{\eea}{\end{eqnarray*}}
\newcommand{\bean}{\begin{eqnarray}}
\newcommand{\eean}{\end{eqnarray}}

\newcommand{\bfth}{\mbox{\boldmath{$\theta$}}}

\newcommand*{\Scale}[2][4]{\scalebox{#1}{$#2$}}%

\DeclareMathOperator*{\argmax}{arg\,max}

\begin{document}

\title{Toward Optimal Feature Selection in Naive Bayes for Text Categorization}

\author{Bo~Tang, ~\IEEEmembership{Student Member,~IEEE,} Steven Kay, ~\IEEEmembership{Fellow,~IEEE,} and Haibo~He, ~\IEEEmembership{Senior Member,~IEEE}
\IEEEcompsocitemizethanks{\IEEEcompsocthanksitem Bo Tang, Steven Kay, and Haibo He are with the Department of Electrical, Computer and Biomedical Engineering at the University of Rhode Island,
Kingston, RI, 02881. E-mail: \{btang, kay, he\}@ele.uri.edu}}

\markboth{Submitted to IEEE Transactions on Knowledge and Data Engineering}%
{}

\IEEEcompsoctitleabstractindextext{
\begin{abstract}
Automated feature selection is important for text categorization to reduce the feature size and to speed up the learning process of classifiers. In this paper, we present a novel and efficient feature selection framework based on the Information Theory, which aims to rank the features with their discriminative capacity for classification. We first revisit two information measures: Kullback-Leibler divergence and Jeffreys divergence for binary hypothesis testing, and analyze their asymptotic properties relating to type I and type II errors of a Bayesian classifier. We then introduce a new divergence measure, called Jeffreys-Multi-Hypothesis (JMH) divergence, to measure multi-distribution divergence for multi-class classification. Based on the JMH-divergence, we develop two efficient feature selection methods, termed maximum discrimination ($MD$) and $MD-\chi^2$ methods, for text categorization. The promising results of extensive experiments demonstrate the effectiveness of the proposed approaches.
\end{abstract}

\begin{keywords}
Feature selection, text categorization, Kullback-Leibler divergence, Jeffreys divergence, Jeffreys-Multi-Hypothesis divergence
\end{keywords}}

\maketitle

\IEEEdisplaynotcompsoctitleabstractindextext

\IEEEpeerreviewmaketitle

\section{Introduction}
\IEEEPARstart{W}{ith} the increasing availability of text documents in electronic form, it is of great importance to label the contents with a predefined set of thematic categories in an automatic way, what is also known as automated \textit{Text Categorization}. In last decades, a growing number of advanced machine learning algorithms have been developed to address this challenging task by formulating it as a classification problem \cite{joachims1998text}\cite{lam1999automatic}\cite{sebastiani2002machine}\cite{al2006new}\cite{aphinyanaphongs2014comprehensive}. Commonly, an automatic text classifier is built with a learning process from a set of prelabeled documents. 

Documents need to be represented in a way that is suitable for a general learning process. The most widely used representation is ``the bag of words": a document is represented by a vector of features, each of which corresponds to a term or a phrase in a vocabulary collected from a particular data set. The value of each feature element represents the importance of the term in the document, according to a specific feature measurement.

A big challenge in text categorization is the learning from high dimensional data. On one hand, tens and hundreds of thousands terms in a document may lead to a high computational burden for the learning process. On the other hand, some irrelevant and redundant features may hurt predictive performance of classifiers for text categorization. To avoid the issue of the ``curse of dimensionality" and to speed up the learning process, it is necessary to perform feature reduction to reduce the size of features. 

A common feature reduction approach for text categorization is feature selection that this paper concentrates on, where only a subset of original features are selected as input to the learning algorithms. In last decades, a number of feature selection methods have been proposed, which can be usually categorized into the following two types of approach: the \textit{filter} approach and the \textit{wrapper} approach \cite{liu2005toward}. The filter approach selects feature subsets based on the general characteristics of the data without involving the learning algorithms that will use the selected features. A score indicating the ``importance" of the term is assigned to each individual feature based on an independent evaluation criterion, such as distance measure, entropy measure, dependency measure and consistency measure. Hence, the filter approach only selects a number of the top ranked features and ignores the rest. Alternatively, the wrapper approach greedily searches for better features with an evaluation criterion based on the same learning algorithm. Although it has been shown that the wrapper approach usually performs better than the filter approach, it has much more computational cost than the filter approach, which sometimes makes it impractical. 

Typically, the filter approach is predominantly used in text categorization because of its simplicity and efficiency. However, the filter approach evaluates the goodness of a feature by only exploiting the intrinsic characteristics of the training data without considering the learning algorithm for discrimination, which may lead to an undesired classification performance. Given a specific learning algorithm, it is hard to select the best filter approach producing the features with which the classifier performs better than all others for discrimination in a viewpoint of theoretical analysis. 

In this paper, we present a feature selection method which ranks the original features, aiming to maximize the discriminative performance for text categorization, when naive Bayes classifiers are used as learning algorithms. Unlike the existing filter approaches, our method evaluates the goodness of a feature without training a classifier explicitly, and selects these features that offer maximum discrimination in terms of a new divergence measure. Specifically, the contributions of this paper are as follows:

\begin{itemize}
\item[1.] We introduce a new divergence measure for multi-class classification by extending the J-divergence measure, termed Jeffreys-Multi-Hypothesis divergence (JMH-divergence). 
\item[2.] We propose an efficient approach to rank the order of features to approximately produce the maximum JMH divergence. The theoretical analysis shows that the JMH divergence is monotonically increasing when more features are selected. 
\item[3.] We analyze the asymptotic distribution of the proposed test statistic, which leads to the $\chi^2$ distribution. By doing so, we introduce another simple and effective feature ranking approach by maximizing the noncentrality measurement of the noncentral $\chi^2$ distribution. 
\end{itemize} 

The rest of the paper is organized as follows: In Section 2, we introduce the previous work on naive Bayes classifiers and feature selection techniques for automatic text categorization. In Section 3, we present the theoretical framework of feature selection using the information measures. In Section 4, we introduce two efficient feature selection approaches in naive Bayes classifiers for text categorization. Experimental results are given in Section 5 along with the performance analysis compared with the state-of-the-art methods. A conclusion and future work discussion are given in Section 6.

\section{Previous Work}
\subsection{Document Representation}
In text categorization, each document is commonly represented by the model of the ``bag-of-words" with a $M \times 1$ feature vector $D : \mathbf{d} = [x_1, x_2, \cdots, x_M]^T$. The $i$-th element $x_i$ corresponds to the measure of the $i$-th term (``word") in a vocabulary or a dictionary. For a given data set, we first generate the vocabulary with a set of $M$ unique terms from all documents. Then, for each document, a feature vector can be formed by using various feature models. Typically, the value of a feature represents the information about this particular term in a document. Two feature models have been widely used. The first one is the binary feature model in which the feature takes value either $0$ or $1$ corresponding to the presence or the absence of a particular term in the document. The distribution of such binary feature for each class can be usually modeled by a Bernoulli distribution. The other one is multi-value feature model in which the feature takes values in $\{0,1,\cdots \}$ corresponding to the number of occurrences of a particular term in the document, and thus it is also called term frequency (TF). The distribution of TF for each class can be usually modeled by a multinomial distribution model. We note here that several other feature models also exist in literature, such as normalized term frequency and inverse document frequency (tf-idf) \cite{salton1988term} and probabilistic structure representation \cite{govert1999probabilistic}. Recent work in learning vector representations of words using neural networks have shown superior performance in classification and clustering \cite{mnih2009scalable}\cite{turian2010word}\cite{mikolov2013efficient}\cite{le2014distributed}, where both the ordering and semantics of the words are considered. 

\subsection{Naive Bayes}
Naive Bayes classifier has been widely used for text categorization due to its simplicity and efficiency \cite{lewis1998naive} \cite{koller1997hierarchically} \cite{li1998classification}. It is a model-based classification method and offers competitive classification performance for text categorization compared with other data-driven classification methods \cite{forman2003extensive}\cite{yang1997comparative} \cite{genkin2007large} \cite{tang2015enn}, such as neural network, support vector machine (SVM), logistic regression, and k-nearest neighbors. The naive Bayes applies the Bayes' theorem with the ``naive" assumption that any pair of features are independent for a given class. The classification decision is made based upon the \textit{maximum-a-posteriori} (MAP) rule. Usually, three distribution models, including Bernoulli model, multinomial model and Poisson model, have commonly been incorporated into the Bayesian framework and have resulted in classifiers of Bernoulli naive Bayes (BNB), multinomial naive Bayes (MNB) and Poisson naive Bayes (PNB), respectively. 

Extensive experiments on real-life benchmarks have shown that the MNB usually outperforms the BNB at large vocabulary size \cite{mccallum1998comparison}. Similar conclusions have been drawn in \cite{sebastiani2002machine} \cite{eyheramendy2003naive} \cite{metsis2006spam}. It has also been shown that the PNB is equivalent to the MNB, if the document length and document class are assumed to be independent \cite{eyheramendy2003naive}. For this reason, the naive Bayes usually refers to the MNB classifier. In this paper, we concentrate on the formulation of the proposed feature selection method for the MNB classifier. The methods can be easily extended to BNB and PNB classifiers.

MNB would be one of the best-known naive Bayes classification approaches using the term frequency to represent the document. Considering a text categorization problem with $N$ classes (``topics"), let $c$ be the discrete variable of class label taking values in $\{1,2,\cdots,N\}$, and $\mathbf{x}$ be the integer-valued feature vector corresponding to the term frequency. The MNB classifier assumes that the number of times that each term appears in the document satisfies a multinomial distribution \cite{lewis1998naive}\cite{mccallum1998comparison}. In other words, a document with $l$ terms is considered as $l$ independent trials, and each term is the result of a trial exactly falling into the vocabulary. Let the vocabulary size be $M$, and then each documents is represented by a $M \times 1$ feature vector.

Hence, given a document $D$, we first count the number of times that each term appears and generate a feature vector $\mathbf{d} = [x_1, x_2, \cdots, x_M]^T$. According to the multinomial distribution, the likelihood of observing $\mathbf{d}$ conditioned on the class label $c$ and the document length $l$ can be calculated as follows:
\begin{align}
\label{mnb_dist}
p(\mathbf{d} | c, l; \bfth^m_c) = \frac{l}{x_1! x_2! \cdots x_M!} \prod_{i=1}^M p_{ic}^{x_i}
\end{align}
where $l = \sum_{m=1}^M x_m$, and $\bfth^m_c = [p_{1c}, p_{2c}, \cdots, p_{Mc}]^T$ is a parameter vector, each of which denotes the probability that the term of a trial falls into one of $M$ categories, so that $p_{ic} \geq 0$ for $i=1, 2, \cdots, M$ and $p_{1c} + p_{2c} + \cdots + p_{Mc} = 1$. Using the MLE method, the estimate of each term probability $\hat{p}_{ic}$ is given by
\begin{align}
\label{mnb_para_mle}
\hat{p}_{ic} = \frac{l_{ic}}{l_c} 
\end{align}
where $l_{ic}$ is the number of times the $i$-th term appears among documents in class $c$, and $l_c$ is the total number of terms in class $c$. To avoid the \textit{zero probability} issue, the technique of ``Laplace smoothing" or the prior information for probability $p_{ic}$ is usually applied \cite{vapnik1982estimation}, which leads to 
\begin{align}
\label{mnb_para}
\hat{p}_{ic} = \frac{l_{ic} + \beta_1}{l_c + \beta_2} 
\end{align}
where $\beta_1$ and $\beta_2$ are the constant smoothing parameters. Using the Laplacean prior, we prime each word's count with a count of one and have $\beta_1 = 1$ and $\beta_2 = M$ \cite{mccallum1998comparison}\cite{eyheramendy2003naive}. 

Note that the document length $l$ is commonly assumed to be independent on the document class to simplify the naive Bayes classification rule, that is, we have $p(\mathbf{d}|c, l) = p(\mathbf{d}|c)$ and $p(c|\mathbf{d},l) = p(c|\mathbf{d})$ for likelihood and posterior probability, respectively. Otherwise, it leads to a more general formulation of the posterior probability for classification, which is given by
\begin{align}
\label{mnb_length}
p(c|\mathbf{d},l) & = p(\mathbf{d},l | c) p(c) \nonumber \\
& = p(\mathbf{d}|c,l) p (l|c) p(c)
\end{align}
where the class-wise document length information $p(l|c)$ is incorporated in a Bayesian fashion. The document length information sometimes may be useful for making classification decisions, e.g., when the class-wise document length distributions are different, but it requires the estimate of $p(l|c)$ for a given data set. We will follow the common assumption that the document length is constant in our experiments, i.e., $p(\mathbf{d}|c) = p(\mathbf{d}|c, l)$, and note that the solutions of our feature selection methods do not change with and without this assumption. 

In naive Bayes classifiers, given a new document $D_t$ to be classified, we first generate its feature vector $\mathbf{d}_t$ and apply the following decision rule to make a classification:
\begin{align}
\label{nb_classifier}
c^{*} & = \argmax_{c \in \{1, 2, \cdots, N \}} p(c|\mathbf{d}_t; {\bfth}_c) \nonumber \\
& \propto \argmax_{c \in \{1, 2, \cdots, N \}} p(\mathbf{d}_t | c; {\bfth}_c) p(c) \nonumber \\
& \propto \argmax_{c \in \{1, 2, \cdots, N \}} \log p(\mathbf{d}_t | c; {\bfth}_c) + \log p(c)
\end{align}
The likelihood probability $p(\mathbf{d}_t | c; {\bfth}_c)$ could be a specific model-based distribution. Here, for MNB classifier, the probability distributions $p(\mathbf{d}_t | c; {\bfth}_c)$ for $i=1,2,\cdots, N$, are the multinomial distributions given by Eq. (\ref{mnb_dist}).

\subsection{Feature Selection}
Feature selection, also called term selection, is a widely adopted approach for dimensionality reduction in text categorization. Given a predetermined integer $r$, the number of terms to be selected, the feature selection approach attempts to select $r$ out of $M$ terms in the original vocabulary. Yang and Pedersen \cite{yang1997comparative} have shown that the employment of feature selection can remove $98\%$ unique terms without hurting the classification performance too much, and thus feature selection can greatly reduce the computational burden for classification. 

In \cite{kohavi1997wrappers}, Kohavi and John have used a so-called \textit{wrapper} feature selection approach in which a feature is either added or removed at one step towards the optimal feature subset selection. When a new feature set is generated, the classifier is re-trained with new features and it is further tested on a validation set. This approach has the advantage of searching the feature space in a greedy way and is always able to find a better feature subset in a sense of an improved classification performance. However, the high computational cost makes it prohibitive for practical text categorization applications.

The alternative and popular one is the \textit{filter} approach, in which each feature is assigned with a score based on its importance measure and only the top ranked features with the highest scores are kept. The advantage of this approach is its easy implementation with low computational cost. In the rest of this section, we describe several state-of-the-art measures that are widely used in text categorization. 

\textit{Document frequency}, the number of documents in which a term occurs is a simple and effective feature selection approach. It removes from the original feature space the rare terms that are considered as non-informative for classification. The effectiveness of this approach also inspired researchers in their experiments to remove all the terms that occur no more than $a$ times ($a$ usually ranges from 1 to 3) in the training set as a preprocessing stage. By doing so, tens of hundreds rare features can be removed before the step of feature selection. \textit{TF-IDF} measure considers both term frequency and inverse document frequency to calculate the importance of features \cite{salton1975vector}. More recently, a generalized TF-IDF measure \cite{bouillot2013extract} is proposed by considering different level of hierarchies among words to effectively analyze tweet user behaviors.

Many other filter approaches are based on the information theory measures, including mutual information, information gain, relevancy score, chi-square statistic, odds ratio, expected cross entropy for text, GSS coefficient, to name a few. We describe some of these measures below. 

\textit{Mutual information} (MI) measures the mutual dependency of two variables. For a term $t_k$ and a category $c_i$, the MI measure between $t_k$ and $c_i$ is defined to be
\begin{align}
MI(t_k, c_i) = \log \frac{p(t_k, c_i)}{p(t_k) p(c_i)} 
\end{align}
where $p(t_k, c_i)$ denotes the probability that the term $t_k$ appears in a document and this document belongs to the category $c_i$, $p(t_k)$ is the probability that the term $t_k$ appears in a document, and $p(c_i)$ is the probability that a document belongs to the category $c_i$. One can see that $MI(t_k, c_i)$ is zero if $t_k$ and $c_i$ is independent, i.e., the term $t_k$ is useless for discriminating the documents belonging to the category $c_i$.   In \cite{mladenic1999feature}, an expected cross entropy for text (CET) is proposed as
\begin{align}
CET(t_k, c_i) = p(t_k, c_i) \log \frac{p(t_k, c_i)}{p(t_k) p(c_i)} 
\end{align}

\textit{Information gain} (IG) \cite{yang1997comparative} \cite{caropreso2001learner} measures the information if one knows the presence or absence of a term in a document, which is defined as 
\begin{align}
IG(t_k, c_i) = p(t_k, c_i) \log \frac{p(t_k, c)}{p(t_k) p(c)} + p(\bar{t}_k, c_i) \log \frac{p(\bar{t}_k, c)}{p(\bar{t}_k) p(c)}
\end{align} 
where $p(\bar{t}_k, c_i)$ denotes the probability that the term $t_k$ does not appear in a document and this document belongs to the category $c_i$, $p(\bar{t}_k)$ is the probability that the term $t_k$ does not appear in a document. Unlike the MI criterion, the IG criterion is less influenced by the low frequency terms and usually performs much better than the MI criterion \cite{yang1997comparative}. 

\textit{Chi-square} statistic is proposed in \cite{yang1997comparative} to measure the lack of independence between the term $t_k$ and the category $c_i$, which is modeled by a Chi-square ($\chi^2$) distribution. By considering the negative evidence of term in a document, a general $\chi^2$ statistic measure is defined as 
\begin{align}
Chi(t_k, c_i) = \frac{\left[ p(t_k, c_i) p(\bar{t}_k, \bar{c}_i) - p(t_k, \bar{c}_i) p(\bar{t}_k, c_i) \right]^2}{p(t_k, c_i) p(t_k, \bar{c}_i) p(\bar{t}_k, c_i) p(\bar{t}_k, \bar{c}_i)}
\end{align} 
where the document space is divided into two categories, $c_i$ and its complement $\bar{c}_i$ that pools all the remaining categories, $p(\bar{t}_k, \bar{c}_i)$ denotes the probability that the term $t_k$ does not appear in a document and also this document does not belong to the category $c_i$, and $p({t}_k, \bar{c}_i)$ denotes the probability that the term $t_k$ appears in a document but this document does not belong to the category $c_i$.

A modified measure termed \textit{GSS coefficient} using negative evidence is proposed by Galavotti et al. in \cite{galavotti2000experiments}, which is defined as
\begin{align}
GSS(t_k, c_i) = p(t_k, c_i) p(\bar{t}_k, \bar{c}_i) - p(t_k, \bar{c}_i) p(\bar{t}_k, c_i)
\end{align}
It has been shown that this measure outperforms the original chi-square measure on several data sets \cite{galavotti2000experiments}.


Notice that almost all of these filter approaches based on the information theory measures use binary variables, e.g., the presence ($t_k$) or the absence $(\bar{t}_k$) of a term in a document, and a document belonging to a category ($c_i$) or not ($\bar{c}_i$). Unlike these existing filter approaches, our proposed approaches make use of the term occurrence to measure the term importance in the document, and hence more richer information is contained. Meanwhile, these existing filter approaches rank the features by only exploring the intrinsic characteristics of data based on the feature relevancy without considering their discriminative information in classifiers. It is  difficult to select an optimal feature subset for discrimination in a theoretical way. In this paper, our new feature selection approaches are able to involve the learning algorithm by maximizing its discriminative capacity. 

\section{Theoretical Framework of Feature Selection}
We follow the Information Theory to select feature subsets that have maximum discriminative capacity for distinguishing the samples among two or more classes. We first introduce some concepts on information measures for binary hypothesis testing (also known as ``two-class" classification) and present a new divergence measure for multiple hypothesis testing (i.e., for ``multi-class" classification).

\subsection{Divergence Measures for Binary Hypothesis Testing}
Considering a two-class classification problem first, each class is represented by a particular distribution, saying $P_1=p(x|c_1;\bfth_1)$ for class $c_1$ and $P_2 = p(x | c_2; \bfth_2)$ for class $c_2$. A test procedure for classification can be considered as a binary hypothesis testing such that if a sample is drawn from $P_1$ we accept the hypothesis $H_1$ (reject the hypothesis $H_2$), and if a sample is drawn from $P_2$ we accept $H_2$ (reject $H_1$). In other words, we have $p(x|c_1) = p(x|H_1)$ and $p(x|c_2) = p(x|H_2)$, and we also denote $p(x|H_i)$ as the class conditional probability distribution in the rest of paper. 

According to the Information Theory \cite{kullback1997information}\cite{cover2012elements}, we define the KL-divergence $\mathcal{KL}(P_1,P_2)$ between two probability distributions (from $P_1$ to $P_2$) as
\begin{align}
\label{KL_12}
\mathcal{KL}(P_1,P_2) & = \int_{x} p(x|H_1) \log \frac{p(x|H_1)}{p(x|H_2)} dx \nonumber \\
& = E_{p_1} \left[ \log \frac{p(x|H_1)}{p(x|H_2)} \right]
\end{align}
where $E_{p_1}[x]$ denotes the expectation of $x$ with respect to the probability distribution $P_1$. Specifically, it is easy to obtain the KL-divergence measure between two discrete distributions that are commonly used in text categorization. 
According to Eq. (\ref{mnb_dist}) and Eq. (\ref{KL_12}), the KL-divergence between two multinomial distributions $P_1^m$ and $P_2^m$ is given by
\begin{align}
\label{KL_MNB}
\mathcal{KL}(P_1^m, P_2^m) = \sum_{i=1}^M p_{i1} \log \frac{p_{i1}}{p_{i2}} 
\end{align}
where $0 \leq p_{ic} \leq 1$ and $\sum_{i=1}^M p_{ic} = 1$ for $i=1,2,\cdots,M$, $c=1,2$. 

Under the MAP rule in Eq. (\ref{nb_classifier}), we would classify the sample $x$ into class $c_1$, i.e., accept $H_1$, if 
\begin{align}
\label{log_lk_ratio}
& \log p(x | H_1) + \log p(H_1) > \log(x | H_2) + p(H_2) \nonumber \\
& \Rightarrow \log \frac{p(x|H_1)}{p(x | H_2)} > -\log \frac{p(H_1)}{p(H_2)} = \gamma
\end{align}
The logarithm of the likelihood ratio, $\log \frac{p(x|H_1)}{p(x | H_2)}$, measures the information of the observation $x$ for discrimination in favor of $H_1$ against $H_2$ \cite{kullback1997information}. From the definition of the KL-divergence measure in Eq. (\ref{KL_12}), then $\mathcal{KL}(P_1,P_2)$ indicates the mean information for discrimination in favor of $H_1$ against $H_2$.

Taking the expectation with respect to the distribution $P_1$ in Eq. (\ref{log_lk_ratio}) (i.e., considering all possible observations drawn from $P_1$), we have 
\begin{align}
\label{KL_12_log}
\mathcal{KL}(P_1, P_2) > \gamma
\end{align}
which also illustrates that $\mathcal{KL}(P_1, P_2)$ is an indicator of a Bayesian classifier's discriminative capacity for discriminating the observation from class $c_1$ in favor of $H_1$ against $H_2$. With the extension of the central limit theorem, Chernoff in \cite{chernoff1952measure} showed that, for a large number of observations, the type I error $\alpha^*$, the probability of incorrectly accepting $H_1$, asymptotically has
\begin{align}
\lim_{n \rightarrow \infty} \log \frac{1}{\alpha^*} = \lim_{n \rightarrow \infty} \mathcal{KL}(P_1, P_2; O_n)
\end{align}
where $O_n$ denotes $n$ independent observations. It has been said that a larger value of KL-divergence indicates a lower type I error when there are infinite number of observations.

Note that the KL-divergence measure is not symmetric. Alternatively, the KL-divergence $\mathcal{KL}(P_2,P_1)$ indicates a Bayesian classifier's discriminative capacity for discriminating the observation from class $c_2$ in favor of $H_2$ against $H_1$. Similarly, under the MAP rule in Eq. (\ref{nb_classifier}), considering all the observation drawn from class $c_2$, we have
\begin{align}
\label{KL_21_log}
\mathcal{KL}(P_2,P_1) > - \gamma
\end{align}
For a large number of observations, the type II error $\beta^*$, the probability of incorrectly accepting $H_2$, can be given by
\begin{align}
\lim_{n \rightarrow \infty} \log \frac{1}{\beta^*} = \lim_{n \rightarrow \infty} \mathcal{KL}(P_2, P_1; O_n)
\end{align}

To minimize both type I error and type II errors in an asymptotic way, the Jeffreys divergence (J-divergence) can be used, which is defined by \cite{jeffreys1946invariant}
\begin{align}
\mathcal{J}(P_1, P_2) = \mathcal{KL}(P_1, P_2) + \mathcal{KL}(P_2, P_1)
\end{align}
By combining Eq. (\ref{KL_12_log}) and Eq. (\ref{KL_21_log}), we have 
\begin{align}
\mathcal{KL}(P_1,P_2) >  \gamma > -\mathcal{KL}(P_2,P_1)
\end{align}
Because
\begin{align}
\mathcal{KL}(P_1,P_2) \geq 0, \quad  \mathcal{KL}(P_2,P_1) \geq 0
\end{align}
a larger $\mathcal{J}(P_1, P_2)$ may lead to a smaller type I and type II error asymptotically. It is known that the J-divergence is able to measure the difficulty and capacity of discriminating between $H_1$ and $H_2$ \cite{kullback1997information} \cite{chernoff1952measure} \cite{jeffreys1946invariant}. 

The purpose of feature selection is to determine the most informative features which lead to the best prediction performance. Hence, it is natural to select those features that have the maximum discriminative capacity for classification, by minimizing the classification error (i.e., 
maximizing the KL-divergence or the J-divergence). However, the J-divergence is only defined for binary hypothesis. We next extend the J-divergence for multiple hypothesis testing (i.e., multi-class classification). 

\subsection{Jeffreys-Multi-Hypothesis Divergence}
The Jensen-Shannon (JS) divergence \cite{lin1991divergence} is the one that can be used to measure multi-distribution divergence, in which the divergences of each individual distribution with a reference distribution are calculated and summed together. Unlike the J-divergence, the measure of discrimination capacity may not hold. In \cite{sawyer1982multiple}, Sawyer presents a variant of J-divergence with a variance-covariance matrix for multiple comparisons of separate hypotheses. Here, we first generalize the J-divergence to a multi-distribution using the scheme of ``one-vs-all" \cite{rifkin2004defense}, which is defined as follows:

\newtheorem{definition}{\bf Definition}
\begin{definition}
\it Let $\mathcal{P}=\{P_1, P_2, \cdots, P_N\}$ be the set of $N$ distributions. The Jeffreys-Multi-Hypothesis (JMH) divergence, denoted by $\mathcal{JMH}(P_1, P_2, \cdots, P_N)$, is defined to be
\begin{align}
\label{JM}
\mathcal{JMH}(P_1, P_2, \cdots, P_N) = \sum_{i=1}^N \mathcal{KL}(P_i, \bar{P}_i)
\end{align}
where $\bar{P}_i$ is the combination of all remaining $N-1$ distributions $\bar{P}_i = \sum_{k = 1,k \neq i}^N \pi_{ki} P_k$, and $\pi_{ki}$ are the prior coefficients. 
\end{definition}

Similar to the ``one-vs-all" strategy for multi-class classification problem, we build $N$ binary hypothesis testing detectors, each of which discriminates the samples in favor of $P_i$ against $\bar{P}_i$ which is the complement of $P_i$. In each detector, $\bar{P}_i$ is represented by a mixture distribution over all the remaining $N-1$ classes with the coefficients $\pi_{ki}$ which are given by
\begin{align}
\pi_{ki} = \frac{p_{c_k}}{\sum_{m=1,m \neq i}^N p_{c_m}}
\end{align}
where $p_{c_m}$ is the prior probability of class $c_m$. Since the KL-divergence of each detector is the measure of its discriminative capacity for discrimination, the new multi-distribution divergence is able to measure the discrimination capacity over all classes. Specifically, when $N=2$, we have $\mathcal{JMH}(P_1, P_2) = \mathcal{J}(P_1, P_2)$. 

Note that, since the JMH divergence is the sum of multiple J-divergences, it holds most properties of J-divergence. For example, JMH divergence is almost positive definite, i.e., $\mathcal{JMH}(P_1, P_2, \cdots, P_N) \geq 0$, with equality if and only if $p_{1m} = p_{2m} = \cdots = p_{Nm}$, $m = 1,2,\cdots,M$. Also it is a symmetric measure, that is, $\mathcal{JMH}(\cdots, P_i, \cdots, P_k, \cdots) = \mathcal{JMH}(\cdots, P_k, \cdots, P_i, \cdots)$.

\section{Selecting the Maximum Discriminative Features}
\subsection{A Greedy Feature Selection Approach}
We consider a binary (two-class) classification problem first and extend our feature selection method to a general multi-class classification problem later. Unlike the existing feature selection methods which compute the score (``importance") of features based on the feature relevance to class, our goal is to select the features that offer the maximum discrimination for classification. By doing so, one can expect an improved classification performance for text categorization. 

For a two-class classification problem, we know that the J-divergence indicates the discriminative capacity of discriminating two classes data under the MAP rule. Hence, we formulate the feature selection problem as follows: given a set of $M$ features $\mathcal{B}$ where $|\mathcal{B}| = M$ and a predetermined integer $r$, the number of features to be selected, we aim to find the most $r$ discriminative features $\mathcal{B}_r^{*} \subset \mathcal{B}$ where $|\mathcal{B}^{*}| = r$, such that,
\begin{align}
\label{feature_selection_problem}
\mathcal{B}_r^{*} & = \argmax_{\mathcal{B}_r \subset \mathcal{B}} \mathcal{J}(P_1, P_2 | \mathcal{B}_r ) \nonumber \\
& = \argmax_{\mathcal{B}_r \subset \mathcal{B}} \mathcal{KL}(P_1, P_2 | \mathcal{B}_r ) + \mathcal{KL}(P_2, P_1 | \mathcal{B}_r )
\end{align}
where $\mathcal{J}(\cdot, \cdot | \mathcal{B}_r )$ and $\mathcal{KL}(\cdot, \cdot | \mathcal{B}_r)$ are the J-divergence and KL-divergence, respectively, when a subset of features $\mathcal{B}_r \subset \mathcal{B}$ are considered. This problem is also known as a NP-hard problem, if one examines each of $\left(\begin{array}{c}
M \\ r
\end{array} \right)$ combinations, which is intractable particularly for a high dimensional data set. Meanwhile, in practice, we need to examine various $r$ values to evaluate the classification performance using those selected $r$ features. Hence, it is necessary to assign an importance score to each feature and rank the features. 

Here, we start to propose a greedy approach to rank the features according to their discriminative capacity for naive Bayes. This approach starts to determine which feature of the $M$ features produces the maximum JMH-divergence if only one single feature is used for classification. To determine the most discrminative feature, for each feature $x_i$, $i=1,2,\cdots, M$, we build two variables: $x_i$ and $\bar{x}_i$, where $x_i$ is the original $i$-th feature and $\bar{x}_i$ is the pool of all the remaining $M-1$ features with parameters $\bar{p}^k_{ic}$. We use the superscript number $k$ to indicate the $k$-th step of our greedy approach (e.g., $k=1$ here). We have 
\begin{align}
\label{remaining_feature}
&\bar{p}^1_{i1} = \sum_{m=1, m \neq i}^M p_{m1} = 1 - p_{i1} \nonumber \\
&\bar{p}^1_{i2} = \sum_{m=1, m \neq i}^M p_{m2} = 1 - p_{i2}
\end{align}
We denote the distributions of these two variables for class 1 and class 2 by $P^1_{i1}$ and $P^1_{i2}$, respectively. Note that $P^1_{i1}$ and $P^1_{i2}$ also satisfy multinomial distribution but with different parameters. Then, we calculate the J-divergence between $P^1_{i1}$ and $P^1_{i2}$ as $\mathcal{J}_i^1(P^1_{i1}, P^1_{i2}) = \mathcal{KL}(P^1_{i1}, P^1_{i2}) + \mathcal{KL}(P^1_{i2}, P^1_{i1})$ with Eq. (\ref{KL_MNB}). At the end of the $1$st step, we obtain $M$ J-divergences $\mathcal{J}_i^1(P^i_{i1}, P^i_{i2})$ for $i=1,2,\cdots, M$, and choose the first feature indexed by $s_1$ that leads to the maximum J-divergence, that is,
\begin{align}
\label{greedy_first_step}
s_1 = \argmax_{i=1,2,\cdots,M} J_i^{1}(P^1_{i1}, P^1_{i2})
\end{align}

Then, we fix the first feature $s_1$ and repeat the process over the remaining features. Specifically, at the $k$-th step to select the $k$-th feature, let $\mathcal{S}_{k-1} = \{s_1, s_2, \cdots, s_{k-1} \}$ be the feature index set which are selected from the previous $k-1$ steps. Again, for each individual feature $x_i$, $i=1,2,\cdots,M$, $i \notin \mathcal{S}_{k-1}$, we form $k+1$ variables: $x_{s_1}, x_{s_2}, \cdots, x_{s_{k-1}}, x_i, \bar{x}_i$, where the first $k$ variables are the original features and the last variable $\bar{x}_i$ is the pool of all remaining $M-k$ features with parameters $\bar{p}^k_{ic}$ for two classes ($c=1,2$). We have 
\begin{align}
\label{remaining_feature}
&\bar{p}^k_{i1} = 1 - p_{i1} - \sum_{m\in\mathcal{S}_{k-1}} p_{m1} \nonumber \\
&\bar{p}^k_{i2} = 1 - p_{i2} - \sum_{m\in\mathcal{S}_{k-1}} p_{m2}
\end{align}
Denote the distributions of these variables for class 1 and class 2 at the $k$-th step by $P^k_{i1}$ and $P^k_{i2}$, respectively. At the $k$-th step, we choose the feature indexed by $s_k$ with the following maximum J-divergences:
\begin{align}
\label{greedy_kstep}
s_k = \argmax_{i=1,2,\cdots,M, i\notin \mathcal{S}_{k-1}} J_i^{k}(P^k_{i1}, P^k_{i2})
\end{align}

Hence, at the end of of the $M$-th step, a ranked feature index set $\mathcal{S}_M = \{s_1, s_2, \cdots, s_M \}$ is produced. The implementation of this greedy feature selection approach based on the maximum J-divergence for two-class classification is given in Algorithm \ref{algorithm: greedy}. 

\begin{figure}[!t]
 \removelatexerror
  \begin{algorithm}[H]
   \label{algorithm: greedy}
   \caption{A Greedy Feature Selection Algorithm Based on the Maximum J-Divergence for Two-class Classification}
   \textbf{INPUT:}
   \begin{itemize}
   \item The estimated probabilities of $M$ features: $\bfth_i = [p_{1c}, p_{2c}, \cdots, p_{Mc}]$, $c=1,2$ for two classes;
   \item The ranked feature index set: $\mathcal{S} = \emptyset$.
   \item The full index set: $\mathcal{S}_a = \{1,2,\cdots,M\}$
   \end{itemize}
   \textbf{ALGORITHM:}\\
      \For{k = 1 : M}{
      \ForEach{i $\in \mathcal{S}_a \cap \bar{\mathcal{S}}$}{
		\textbf{\textit{1.}} Form $k+1$ variables: $x_{s_1}, x_{s_2}, \cdots$, $x_{s_{k-1}}$, $x_i$ and $\bar{x}_i$, and denote $P^k_{i1}$ and $P^k_{i2}$ for class 1 and 2 distribution, respectively;\\
		\textbf{\textit{2.}} Calculate J-divergence $J_i^{k}(P^k_{i1}, P^k_{i2})$ between $P^k_{i1}$ and $P^k_{i2}$;\\		
      }
	\textbf{\textit{3.}} $s_k$: select the feature using Eq. (\ref{greedy_kstep});\\
	\textbf{\textit{4.}} $\mathcal{S} = \mathcal{S} \cup s_k$;\\ 
    }
    \textbf{OUTPUT:}  \\
    \begin{itemize}
    \item A ranked feature index set: $\mathcal{S} = \{s_1, s_2, \cdots, s_M \}$.
    \end{itemize}
  \end{algorithm}
\end{figure}

\newtheorem{theorem}{\bf Theorem}
\begin{theorem}
\label{theorem1}
\it The maximum J-divergences $\mathcal{J}_{s_k}$ for $k=1,2,\cdots, M$ in Algorithm 1 monotonically increases, i.e.,
\begin{align}
\mathcal{J}^{k+1}_{s_{k+1}} (P^{k+1}_{s_{k+1}1}, P^{k+1}_{s_{k+1}2}) \geq \mathcal{J}^k_{s_{k}} (P^{k}_{s_{k}1}, P^{k}_{s_{k}2})
\end{align}
\end{theorem}

The proof of Theorem 1 is provided in our Supplemental Material. This theorem indicates that the discriminative capacity increases when more features are used for classification, under the assumption that the term occurrence of a document satisfies a particular multinomial distribution. 
Note that the proposed greedy feature selection algorithm makes a locally optimal choice at each step to approximate the global optimal solution of Eq. (\ref{feature_selection_problem}), by selecting a feature with the maximum discriminative capacity for classification. The significance of this algorithm is that it starts at the best first feature and towards the optimal solution. 

This greedy approach can be considered as a wrapper approach. However, unlike the existing wrapper approaches, this greedy approach does not need to evaluate the classification performance on a validation data set through retraining the classifier when a new feature is generated, because a closed form of KL-divergence is given in Eq. (\ref{KL_MNB}) to measure the discriminative capacity of MNB classifiers. 

However, this greedy approach still has the computational complexity of $\mathcal{O}(M^2/2)$, which leads to a heavy computational load for a high-dimensional data set. Next, we provide a more efficient feature selection approach for text categorization. 

\subsection{An Efficient Feature Selection Approach}
In Algorithm \ref{algorithm: greedy}, the best one single feature is selected at the first step, providing an optimal starting point to approximate the optimal solution. At this step, we rank the J-divergences over all features, which are given by
\begin{align}
\label{efficient_first_step}
\Scale[0.9]{J^1_{e_1}(P^1_{e_1 1}, P^1_{e_1 2}) \geq J^1_{e_2}(P^1_{e_2 1}, P^1_{e_2 2}) \geq \cdots \geq J^1_{e_M}(P^1_{e_M 1}, P^1_{e_M 2})}
\end{align}
where $e_i$ denotes the feature index and we know $e_1$ is $s_1$ which is given by Eq. (\ref{greedy_first_step}) in Algorithm \ref{algorithm: greedy}. Looking at the first two J-divergences in Eq. (\ref{efficient_first_step}), we have
\begin{align}
\label{efficient_first_step_2}
&p_{e_1 1} \log \frac{p_{e_1 1}}{p_{e_1 2}} + (1 - p_{e_1 1}) \log\frac{1-p_{e_1 1}}{1-p_{e_1 2}} \nonumber \\
& \geq p_{e_2 1} \log \frac{p_{e_2 1}}{p_{e_2 2}} + (1 - p_{e_2 1}) \log\frac{1-p_{e_2 1}}{1-p_{e_2 2}}
\end{align}
Since one single term probability is very small in practice, i.e., $p_{i1} \ll 1$ for $i=1,2,\cdots, M$, the term $\log\frac{1-p_{i1}}{1-p_{i2}}$ would be very close to zeros, and then Eq. (\ref{efficient_first_step_2}) may lead to 
\begin{align}
\label{efficent_first_ineq}
p_{e_11} \log \frac{p_{e_1 1}}{p_{e_1 2}} \geq p_{e_2 1} \log \frac{p_{e_2 1}}{p_{e_2 2}}
\end{align}

At the second step in Algorithm \ref{algorithm: greedy}, the feature $e_2$ is usually selected due to the fact that $\log\frac{1-p_{e_1 1} - p_{e_2 1} }{1-p_{e_1 2} - p_{e2 2}} \approx 0$. That is because, according to Eq. (\ref{efficent_first_ineq}), we have
\begin{align}
\label{efficient_approx}
p_{e_1 1} \log \frac{p_{e_1 1}}{p_{e_1 2}} + p_{e_2 1} \log \frac{p_{e_2 1}}{p_{e_2 2}} \geq p_{e_1 1} \log \frac{p_{e_1 1}}{p_{e_1 2}} + p_{j 1} \log \frac{p_{j 1}}{p_{j 2}}
\end{align}
for $j=1,2,\cdots, M$, $j \neq e_1, e_2$. 

Therefore, instead of doing a greedy search, an efficient way is to use the ranked feature index set $\mathcal{E} = \{e_1, e_2, \cdots, e_M\}$ in Eq. (\ref{efficient_first_step}). We summarize this efficient feature selection algorithm in Algorithm \ref{algorithm: two_class_efficient}. Compared to the Algorithm \ref{algorithm: greedy}, the proposed approach is much more \textit{efficient} as each feature score is calculated once, and it has the computational complexity of $\mathcal{O}(M)$. This efficient approach evaluates the ``importance" of each individual feature by measuring its discriminative capacity, when only one single feature is used for classification. We note that the Theorem \ref{theorem1} is also satisfied for this efficient approach, i.e., the J-divergence measure increases as more features are selected. Meanwhile, we also note that the features selected in Algorithm 2, are not necessarily the same as ones selected in Algorithm 1, as one can see, for example, that Eq. (\ref{efficent_first_ineq}) and Eq. (\ref{efficient_approx}) approximately hold for the first and second feature selection. Considering the first feature selection as an example, this approximation depends on the value of $\Delta$ defined as
\begin{align}
\Delta  = (1 - p_{e_2 1}) \log \frac{1 - p_{e_2 1} }{1 - p_{e_2 2} }  - & (1 - p_{e_1 1})  \log \frac{1 - p_{e_1 1}}{1 - p_{e_1 2} }
\end{align}
More precisely, given $e_1$ is selected as the first feature in Algorithm 2, $e_1$ also ranks first in Algorithm 1 if and only if the following condition holds:
\begin{align}
\label{sufficient_condition}
\Delta \leq p_{e_11} \log \frac{p_{e_1 1}}{p_{e_1 2}} - p_{e_2 1} \log \frac{p_{e_2 1}}{p_{e_2 2}}
\end{align}
%
%


\begin{figure}[!t]
 \removelatexerror
  \begin{algorithm}[H]
   \label{algorithm: two_class_efficient}
   \caption{An Efficient Feature Selection Algorithm Based on the Maximum J-Divergence for Two-class Classification}
   \textbf{INPUT:}
   \begin{itemize}
   \item The estimated probabilities of $M$ features: $\bfth_i = [p_{1c}, p_{2c}, \cdots, p_{Mc}]$, $c=1,2$ for two classes;
   \end{itemize}
   \textbf{ALGORITHM:}\\
   \For{i = 1 : M}{
  		\textbf{\textit{1.}} Form two variables: $x_{i}$ and $\bar{x}_i$, and denote $P_{i1}$ and $P_{i2}$ for class 1 and 2 distribution, respectively;\\
		\textbf{\textit{2.}} Calculate J-divergence $J_i(P_{i1}, P_{i2})$ between $P_{i1}$ and $P_{i2}$ using Eq. (\ref{KL_MNB}) as score for the $i$-th feature;\\		
    }   
	\textbf{\textit{3.}} Sort the feature scores in a descend way: $\mathcal{J}_{e_1} > \mathcal{J}_{e_2} > \cdots > \mathcal{J}_{e_M}$.\\
   \textbf{OUTPUT:}
   \begin{itemize}
	\item  A ranked feature index set: $\mathcal{E} = \{ e_1, e_2, \cdots, e_M \}$.   
   \end{itemize}
  \end{algorithm}
\end{figure}

\subsection{Multi-class Classification}
In this section, we extend the above efficient feature selection method for multi-class classification problems. Considering an $N$-class classification problem, the $r$ most discriminative features $\mathcal{B}_r^{*}$ are selected by maximizing the JMH-divergence, which are given by
\begin{align}
\label{feature_selection_problem_multiple}
\mathcal{B}_r^{*} &= \argmax_{\mathcal{S}_r \subset \mathcal{B}, |\mathcal{S}_r| = r} \mathcal{JMH}(P_1, P_2, \cdots, P_N | \mathcal{S}_r ) \nonumber \\
&=\argmax_{\mathcal{S}_r \subset \mathcal{B}, |\mathcal{S}_r| = r} \sum_{i=1}^N \mathcal{KL}(P_i, \bar{P}_i | \mathcal{S}_r)
\end{align}
where $\mathcal{JMH}(P_1, P_2, \cdots, P_N | \mathcal{S}_r )$ is the JMH-divergence defined in Eq. (\ref{JM}) with the feature subset $\mathcal{S}_r$. Note that each KL-divergence indicates the discriminative capacity of one binary classifier to distinguish the samples in one class $P_i$ from the samples in all remaining classes $\bar{P}_i$, and thus the JMH-divergence is able to measure the difficulty and capacity of discriminating the samples among all classes. 

The efficient feature selection method for an $N$-class classification problem based on the maximum JMH-divergence is presented in Algorithm \ref{algorithm: multi_class_efficient}. The value of JMH-divergence is used as the score for each feature. Sorting the feature scores in a descend way, we output a ranked feature index set $\mathcal{E} = \{ e_1, e_2, \cdots, e_M \}$ for multi-class classification. The computational complexity of this algorithm is $\mathcal{O}(MN)$.

\begin{figure}[!t]
 \removelatexerror
  \begin{algorithm}[H]
   \label{algorithm: multi_class_efficient}
   \caption{An Efficient Feature Selection Algorithm Based on the Maximum JMH-Divergence for $N$-class Classification}
   \textbf{INPUT:}
   \begin{itemize}
   \item The estimated probabilities of $M$ features: $\bfth_c = [p_{1c}, p_{2c}, \cdots, p_{Mc}]$, $c=1,2, \cdots, N$;\\
   \item The prior probabilities of $N$ classes: $p_{1}, p_{2}, \cdots, p_{N}$;\\
   \end{itemize}
   \textbf{ALGORITHM:}\\
   \For{i = 1 : M}{
   		\textbf{\textit{1.}} Form two variables: $x_{i}$ and $\bar{x}_i$, and denote $P_{ic}$ as the class distribution, $c=1,2,\cdots,N$;\\	
   		\For{c = 1 : N}{
   			\textbf{\textit{2.}} Form two distributions: $P_{ic}$ and $\bar{P}_{ic}$, where $\bar{P}_{ic}$ is the one grouping all remaining $N-1$ classes;\\	
   			\textbf{\textit{3.}} Calculate KL-divergence $\mathcal{KL}(P_{ic}, \bar{P}_{ic})$ between $P_{ic}$ and $\bar{P}_{ic}$;\\
   		}
   		\textbf{\textit{4.}} Calculate the JMH-divergence $\mathcal{JMH}_i = \sum_{c=1}^N \mathcal{KL}(P_{ic}, \bar{P}_{ic})$ as the score for the $i$-th feature;\\		
    }   
	\textbf{\textit{5.}} Sort the feature scores in a descend way: $\mathcal{JMH}_{e_1} > \mathcal{JMH}_{e_2} > \cdots > \mathcal{JMH}_{e_M}$.\\
   \textbf{OUTPUT:}
   \begin{itemize}
	\item A ranked feature index set: $\mathcal{E} = \{ e_1, e_2, \cdots, e_M \}$.   
   \end{itemize}
  \end{algorithm}
\end{figure}

\subsection{Feature Selection Based on $\chi^2$ Statistics}
The KL-divergence measure $\mathcal{KL}(P_1, P_2)$ is also known as the minimum discrimination information between the probability distributions $P_1$ and $P_2$. Suppose that we have a random sample of $l$ observations (e.g., a document with length $l$), and we try to test a null hypothesis $H_2$, the observation is drawn from class 2 with the distribution $P_2$, against an alternative hypothesis $H_1$, the observation is drawn from class 1 with the distribution $P_1$. The minimum discrimination statistic \cite{kullback1997information} in favor of $H_1$ against $H_2$ is defined as
\begin{align}
\mathcal{D}(\hat{P}_1, P_2) = \hat{P}_1 \log \frac{\hat{P}_1}{P_2}
\end{align}
where $\hat{P}_1$ is the estimate of $P_1$ from the given observations and $P_2$ is assumed to be known. We may reject the null hypothesis $H_2$ and accept the alternative hypothesis $H_1$ if the value of statistic $\mathcal{D}(\hat{P}_1, P_2)$ exceeds a predetermined threshold. Asymptotically, the statistic $2\mathcal{D}(\hat{P}_1, P_2)$ under the null hypothesis $H_2$ satisfies a central Chi-squared distribution $\chi^2_{M-1}$ with $M-1$ degrees of freedom, and satisfies a non-central Chi-squared distribution $\chi^2_{M-1}(\nu)$ under the alternative hypothesis $H_1$. The noncentrality parameter $\nu_D$ is given by
\begin{align}
\nu_D(\hat{P}_1,P_2) = l \sum_{i=1}^M \frac{( \hat{p}_{i1} - p_{i2})^2 }{p_{i2}}
\end{align}
where $\hat{p}_{i1}$ is the estimate of $p_{i1}$ under $H_1$ using the MLE method from the given observations. Asymptotically, the J-divergence $\mathcal{J}(\hat{P}_1, P_2)$ is the sum of two $\chi^2$ distributions. We have
\begin{align}
\label{chi_J}
\nu_J(\hat{P}_1,P_2) = \frac{l}{2} \sum_{i=1}^M \frac{( \hat{p}_{i1} - p_{i2})^2 }{p_{i2}} + \frac{l}{2} \sum_{i=1}^M \frac{( \hat{p}_{i1} - p_{i2})^2 }{\hat{p}_{i1}}
\end{align}
where the former one is known as the Pearson's $\chi^2$ and the latter one is also known as the Neyman's $\chi^2$ \cite{kullback1997information}. 

Thus, one can see that the noncentrality parameter $\nu_J$ would also be a good sign to indicate the discriminative capacity of discriminating between $H_1$ and $H_2$. Therefore, we can further select the features by maximizing the noncentrality parameter $\nu_J$ for binary classification. Under the assumption that the number of samples in training data set goes infinity, we use the estimation of $p_{i1}$ and $p_{i2}$ for $i=1,2,\cdots, M$ from training data in Eq. (\ref{chi_J}).  For a multi-class classification problem, unlike the Algorithm \ref{algorithm: multi_class_efficient}, each feature is assigned with a score $\mathcal{CHI}_i = \sum_{c=1}^N \nu(P_{ic}, \bar{P}_{ic})$ which is the sum of $N$ noncentrality parameters in their $\chi^2$ distributions, and the ranked feature index set $\mathcal{S} = \{e_1, e_2, \cdots, e_M\}$ is produced by sorting the scores: 
\begin{align}
\label{chi-squared test}
\mathcal{CHI}_{e_1} > \mathcal{CHI}_{e_2} > \cdots > \mathcal{CHI}_{e_M}
\end{align}

We need to note here that the feature selection approach based upon the Chi-squared statistic in Eq. (\ref{chi-squared test}) should be equivalent to the approach in Algorithm \ref{algorithm: multi_class_efficient} if there are infinite training documents. When the assumption of large numbers is not satisfied, the features selected by Eq. (\ref{chi-squared test}) may lose some discriminative capacity. As we demonstrated through extensive experiments, the discrimination performance of the Chi-squared statistic is usually bounded by the approach in Algorithm \ref{algorithm: multi_class_efficient}. However, we sometimes also observe that the Chi-squared statistic performs better for some real-life text data sets. Therefore, in practice, we would like to recommend the use of both approaches because of their simplicity, efficiency and improved discrimination performance.

\section{Experimental Results and Analysis}

\subsection{Real-Life Data Sets}
For our experiments,\footnote{We also verify the effectiveness of the proposed approaches for synthetic data sets. The detailed simulation results and analysis are given in our Supplemental Material.} we test our proposed feature selection approaches on three benchmarks that have been prepared by Deng et al. \cite{cai2005document} \cite{cai2008modeling} for text categorization: \textsc{20-Newsgroups}, \textsc{Reuters}, and \textsc{Topic Detection and Tracking (TDT2)}. These three benchmarks have been widely used in literature for performance evaluation. The \textsc{20-Newsgroups} benchmark consists of about $20,000$ documents collected from the postings of $20$ different online newsgroups or topics.

The \textsc{Reuters} originally contains $21,578$ documents with $135$ topics, but some documents belong to multiple topics. For our experiments, we use the ModApte version of the Reuters by removing those documents with multiple labels. This version consists of $8,293$ documents in $65$ topics. Following the work in \cite{mccallum1998comparison}, we form three data sets, named \textsc{Reuters-10}, \textsc{Reuters-20} and \textsc{Reuters-30}, consisting of the documents of the first 10, 20 and 30 topics, respectively.

The \textsc{TDT2} benchmark consists of $11,201$ documents taken from two newswires (AP WorldStream and New York Times Newservice), two radio programs (PRI The World and VOA World News) and two television programs (CNN Headline News and ABC World News Tonight). Also, those documents that belong two or more topics have been removed. Because of the extremely imbalanced data for some categories, we only use the first 10 topics with the largest data size as our data set. 

For all these data sets used in our experiments, we ignore those words in a stoplist, and discard those words that appear in less than 2 documents or messages in our preprocessing stage. For all data sets except \textsc{TDT2}, we perform classification on the officially split training and testing data sets. For the TDT2, we use 10-fold cross validation for performance evaluation, and the reported results are averaged over 10 runs. 

\subsection{Performance Evaluation Metrics}
We use the following metrics to evaluate the classification performance: \textit{accuracy}, \textit{precision}, \textit{recall}, and \textit{F1 measure}. The accuracy metric is widely used in machine learning fields, which indicates the overall classification performance. The precision is the percentage of documents that are correctly classified as positive out of all the documents that are classified as positive, and the recall is the percentage of documents that are correctly classified as positive out of all the documents that are actually positive. The metrics of precision and recall are defined as
\begin{align}
\textit{Precision} = \frac{\textit{TP}}{\textit{TP} + \textit{FP}} \nonumber \\
\textit{Recall} = \frac{\textit{TP}}{\textit{TP} + \textit{FN}}
\end{align}
where \textit{TP} denotes the number of true positive, \textit{FP} denotes the number of false positive, and \textit{FN} denotes the number of false negative. These two metrics have an inverse relationship between each other. In other words, increasing the precision is at the cost of reducing the recall, and vice versa. Among those measures that attempt to combine precision and recall as one single measure, the F1 measure is one of the most popular, which is defined by
\begin{align}
\textit{F1} = \frac{2 \times \textit{Precision} \times \textit{Recall}}{\textit{Precision} + \textit{Recall}}
\end{align}
The metrics of precision, recall and F1 measure are originally defined for binary class. For multi-class classification, we follow several other studies \cite{joachims1998text}\cite{mccallum1998comparison}\cite{liere1997active}\cite{combarro2005introducing}, in which binary classifiers are built for each individual class and a global F1 measure is obtained by averaging the F1 measure of each class weighted by the class prior. 

\subsection{Results}
We compare our two efficient feature selection approaches: the maximum discrimination termed \textit{MD} and its asymptotic $\chi^2$ statistic termed \textit{MD-$\chi^2$}, with the state-of-the-art feature ranking methods, including document frequency (\textit{DF}), expected cross entropy for text (\textit{CET}), $\chi^2$ statistic and \textit{GSS}. We carry out experiments on these three benchmarks when naive Bayes and SVM are used as classifiers. To compare the performance of these feature selection methods, we evaluate the classification accuracy and the F1 measure metric of these classifiers with different number of features ranging from $10$ to $2,000$. 

\begin{figure}[H]
\centering
\includegraphics[width=.7\linewidth]{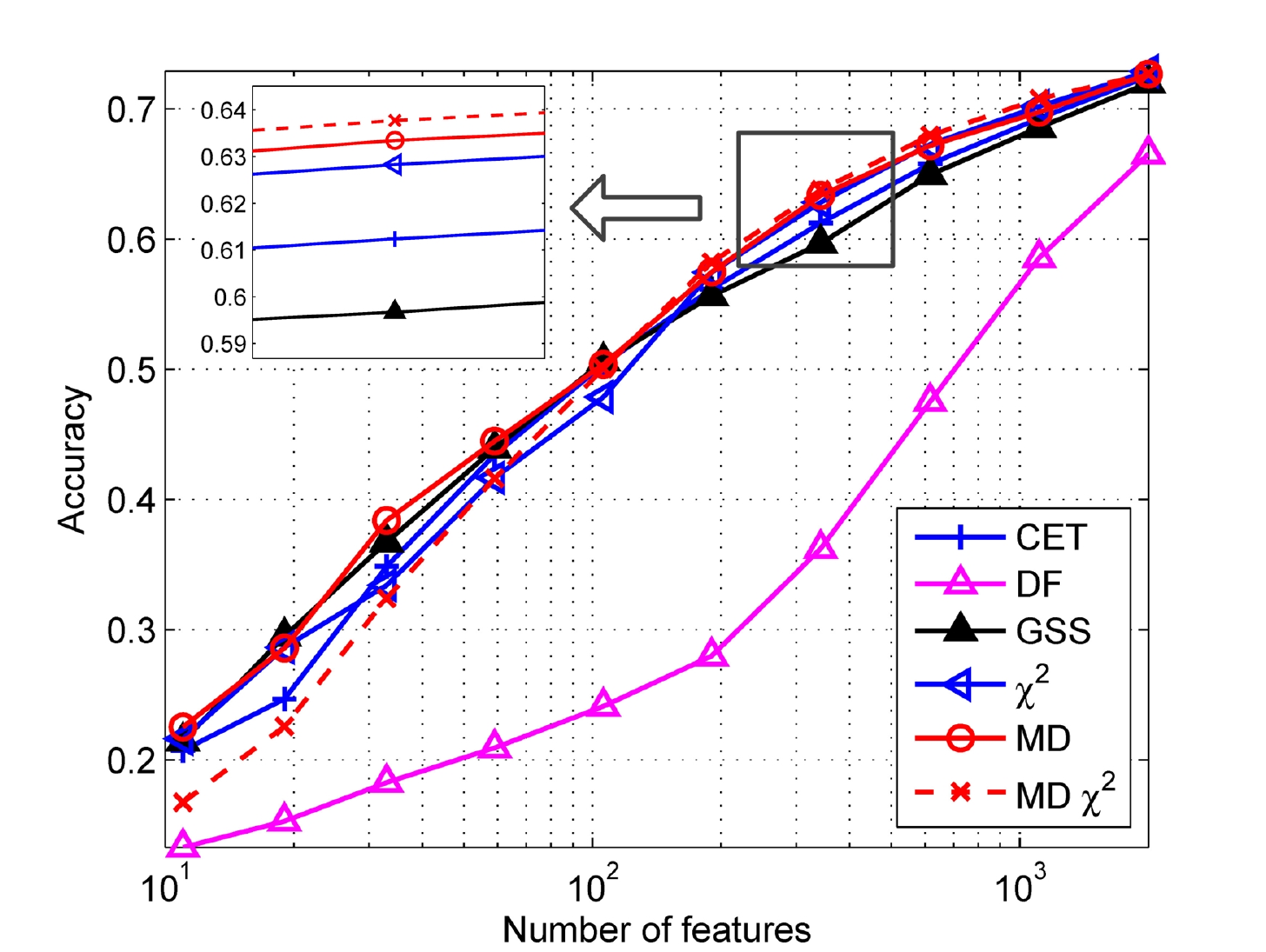}
\caption{An accuracy comparison of feature selection methods on the \textsc{20-Newsgroups} with $20$ topics for naive Bayes.}
\label{20news_all}
\end{figure}

\begin{figure}[htp]
\captionsetup[subfigure]{labelformat=empty}
  \centering
  \subcaptionbox{(a). Accuracy for \textsc{alt-comp} \label{alt_comp:a}}{\includegraphics[width=2.4in]{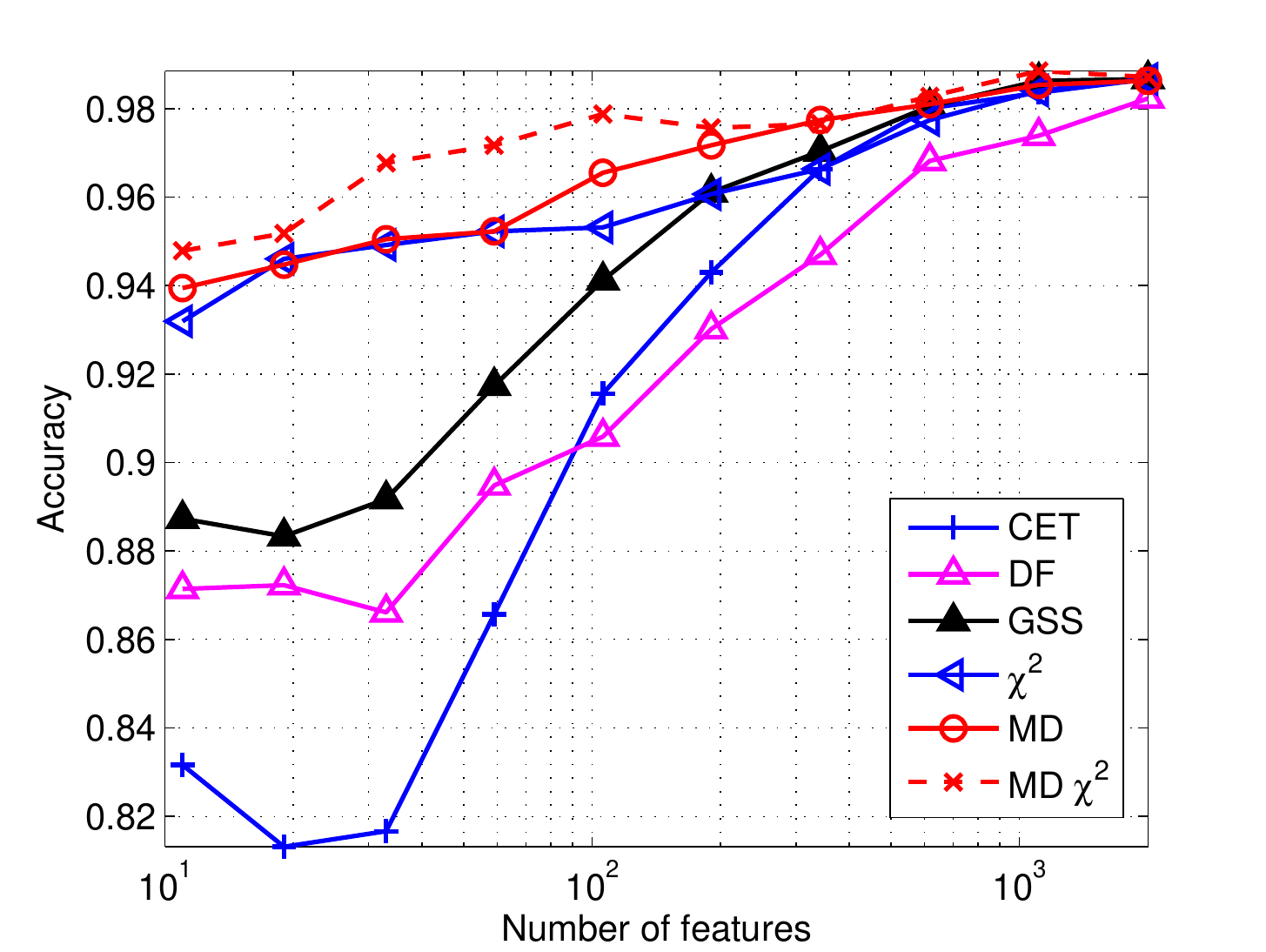}}
  \qquad
  \subcaptionbox{(b). F1 measure for \textsc{alt-comp} \label{alt_comp:b}}{\includegraphics[width=2.4in]{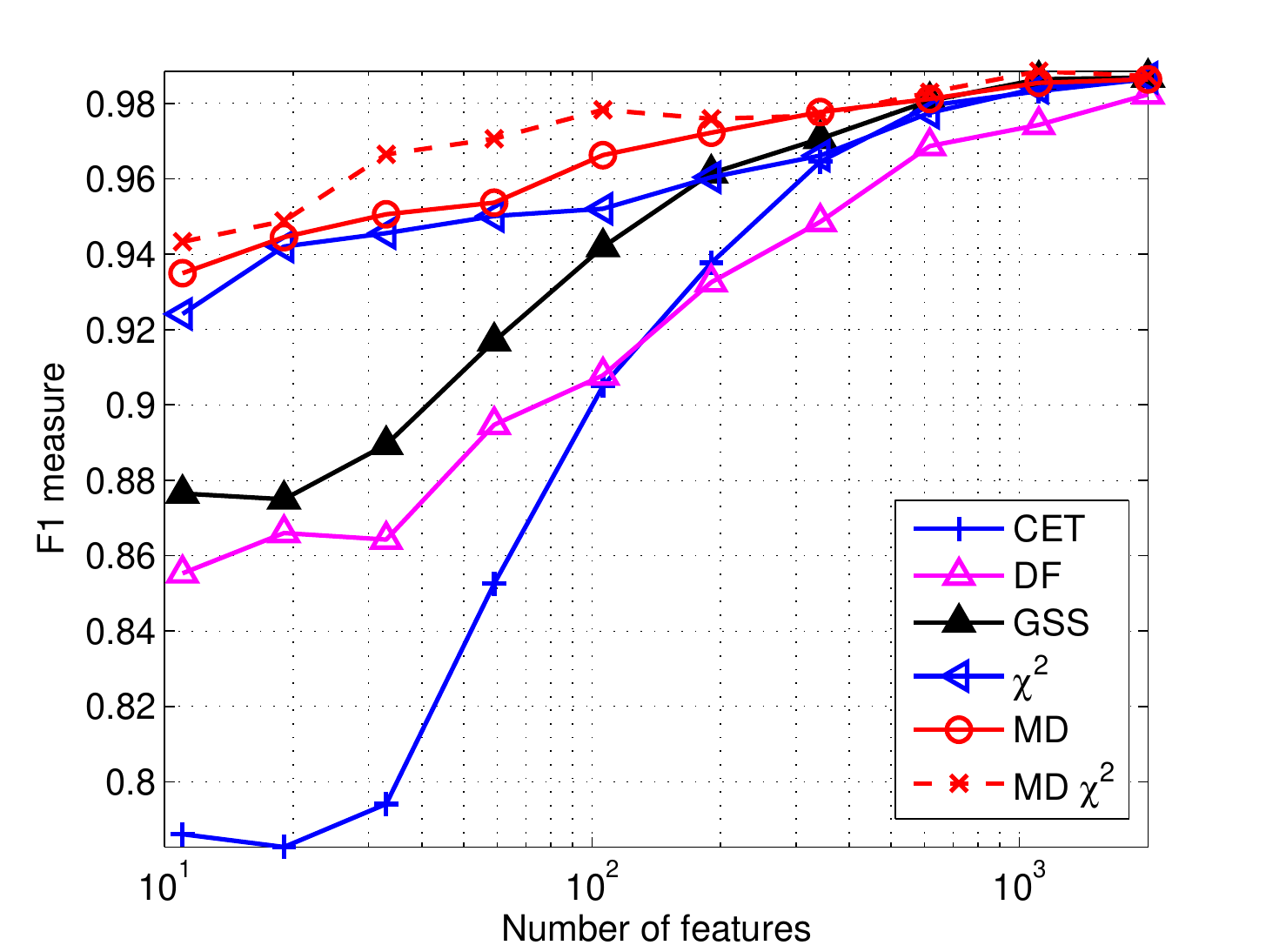}}\hspace{0.1em}%
  \caption{Performance comparisons of feature selection methods on the \textsc{alt-comp} data set: (a) accuracy, (b) F1 measure, when naive Bayes is used as the classifier. }
  \label{alt_comp}
\end{figure}

\begin{figure*}[htp]
\captionsetup[subfigure]{labelformat=empty}
  \centering
  \subcaptionbox{(a-1). Accuracy for \textsc{Reuters-10} \label{Reuters:a1}}{\includegraphics[width=2.3in]{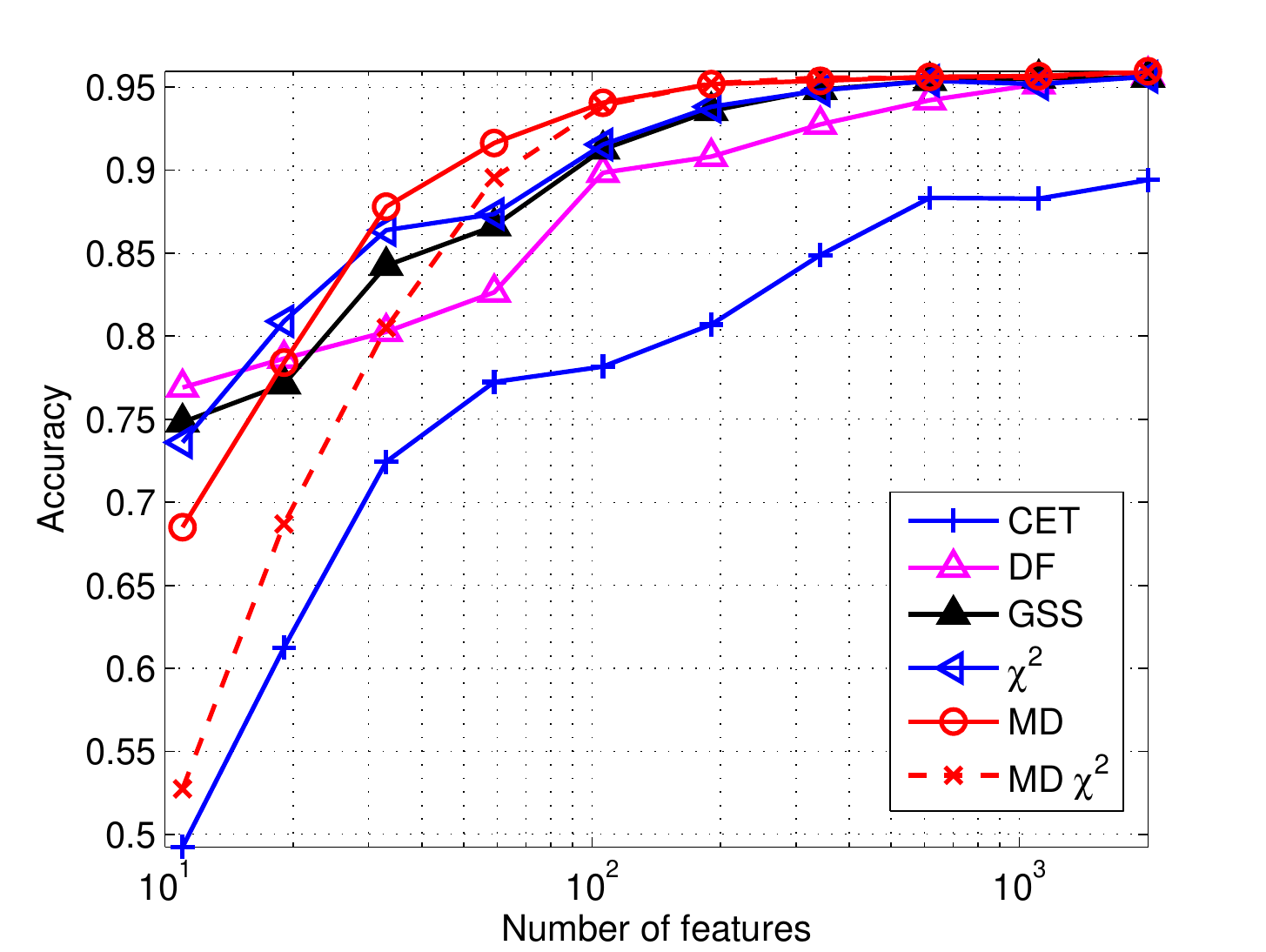}}\hspace{0.1mm}%
  \subcaptionbox{(a-2). F1 measure for \textsc{Reuters-10} \label{Reuters:a2}}{\includegraphics[width=2.3in]{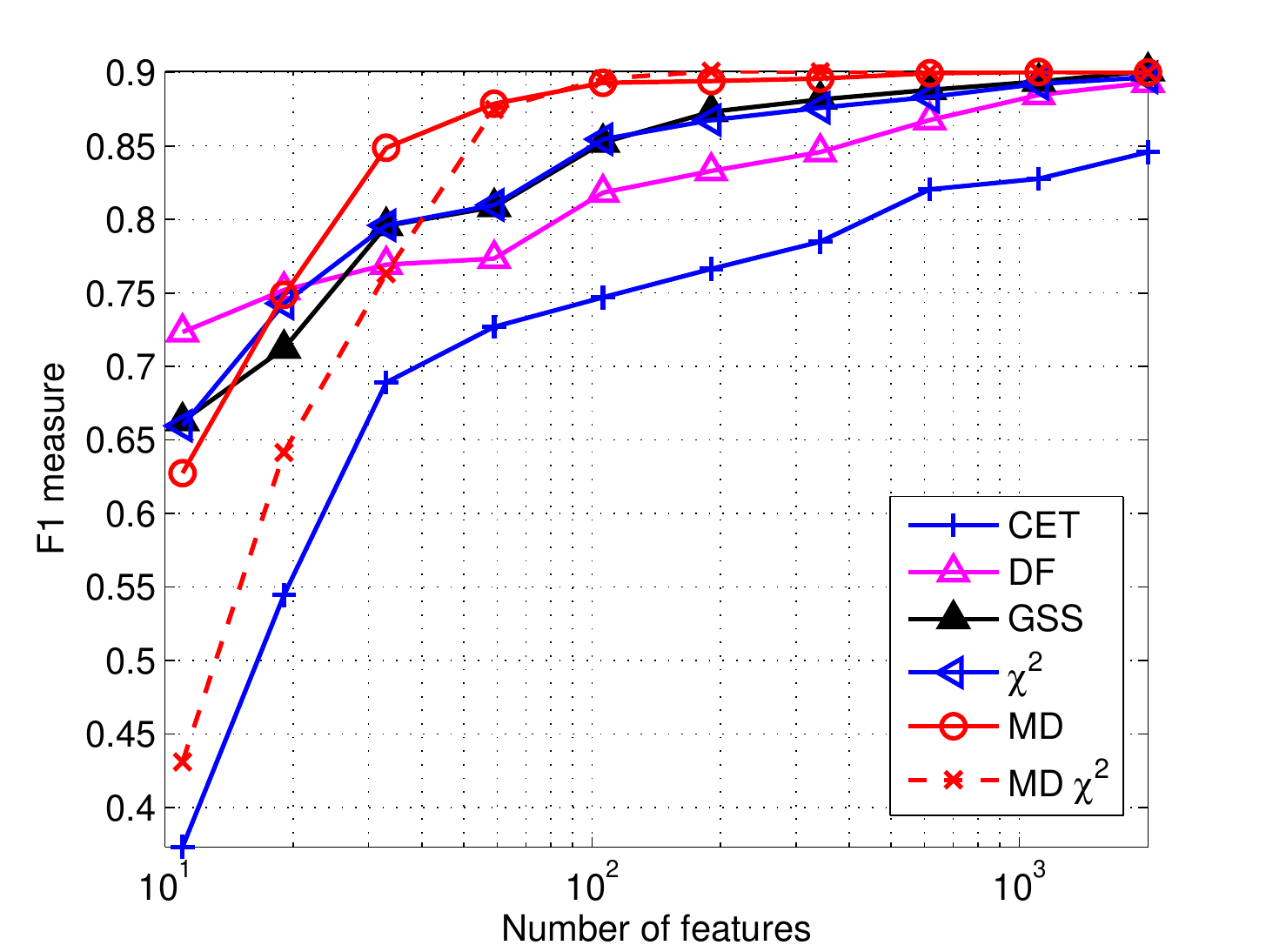}}\hspace{0.1em}%
  
  \subcaptionbox{(b-1). Accuracy for \textsc{Reuters-20} \label{Reuters:b1}}{\includegraphics[width=2.3in]{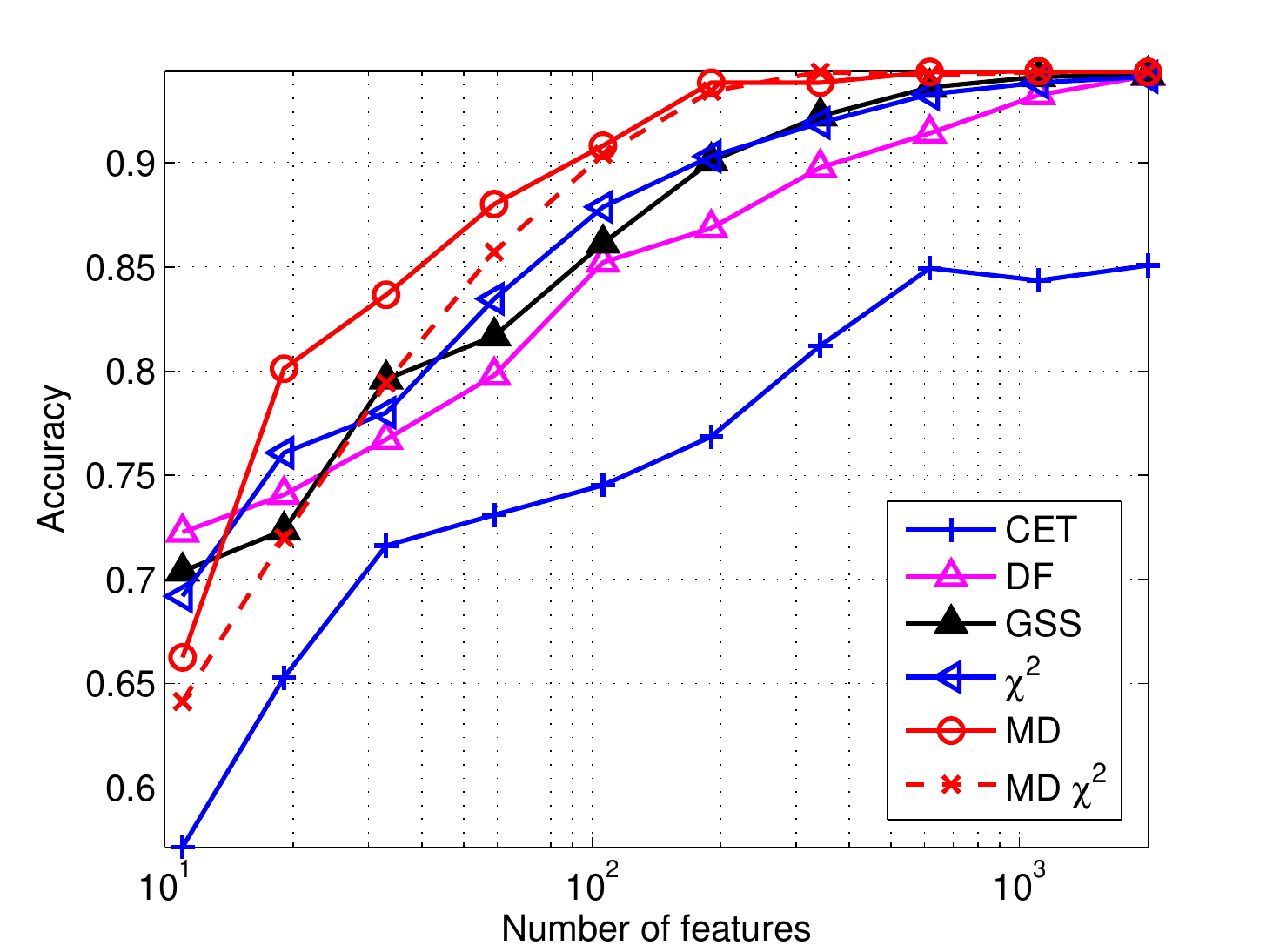}}\hspace{0.1em}%
  \subcaptionbox{(b-2). F1 measure for \textsc{Reuters-20}  \label{Reuters:b2}}{\includegraphics[width=2.3in]{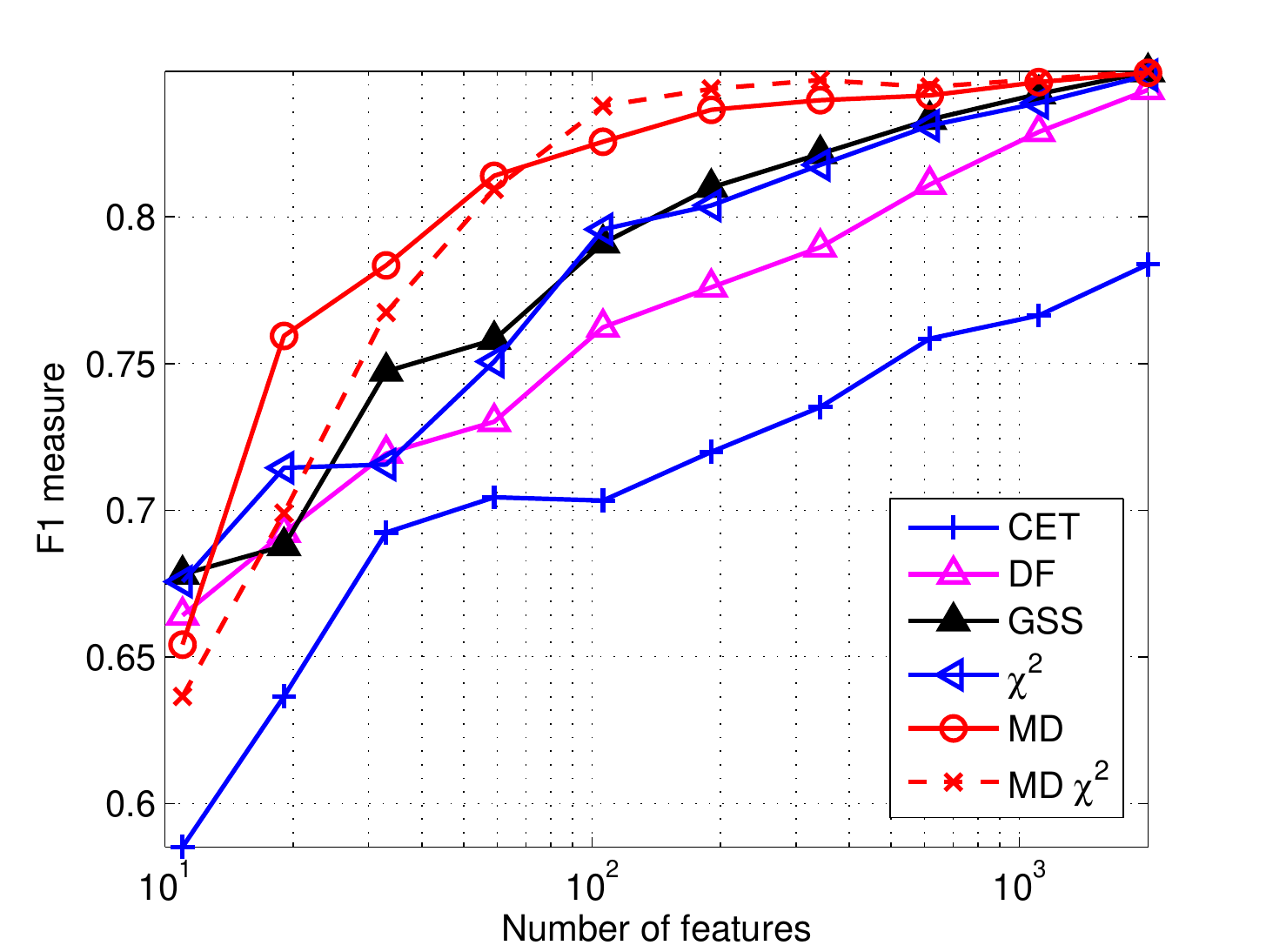}}
  
  \subcaptionbox{(c-1). Accuracy for \textsc{Reuters-30} \label{Reuters:c1}}{\includegraphics[width=2.3in]{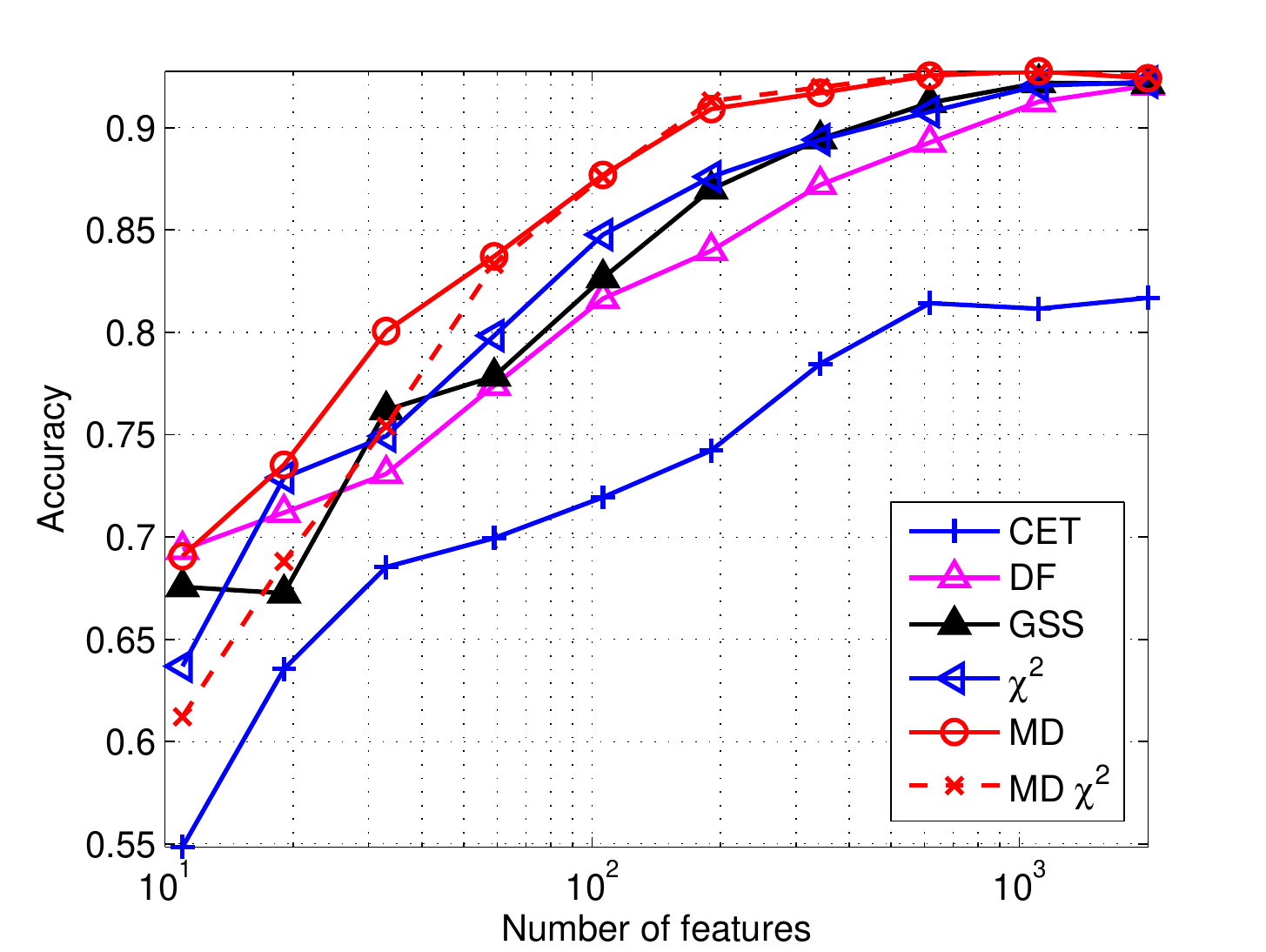}}\hspace{0.1em}%
  \subcaptionbox{(c-3). F1 measure for \textsc{Reuters-30} \label{Reuters:c2}}{\includegraphics[width=2.3in]{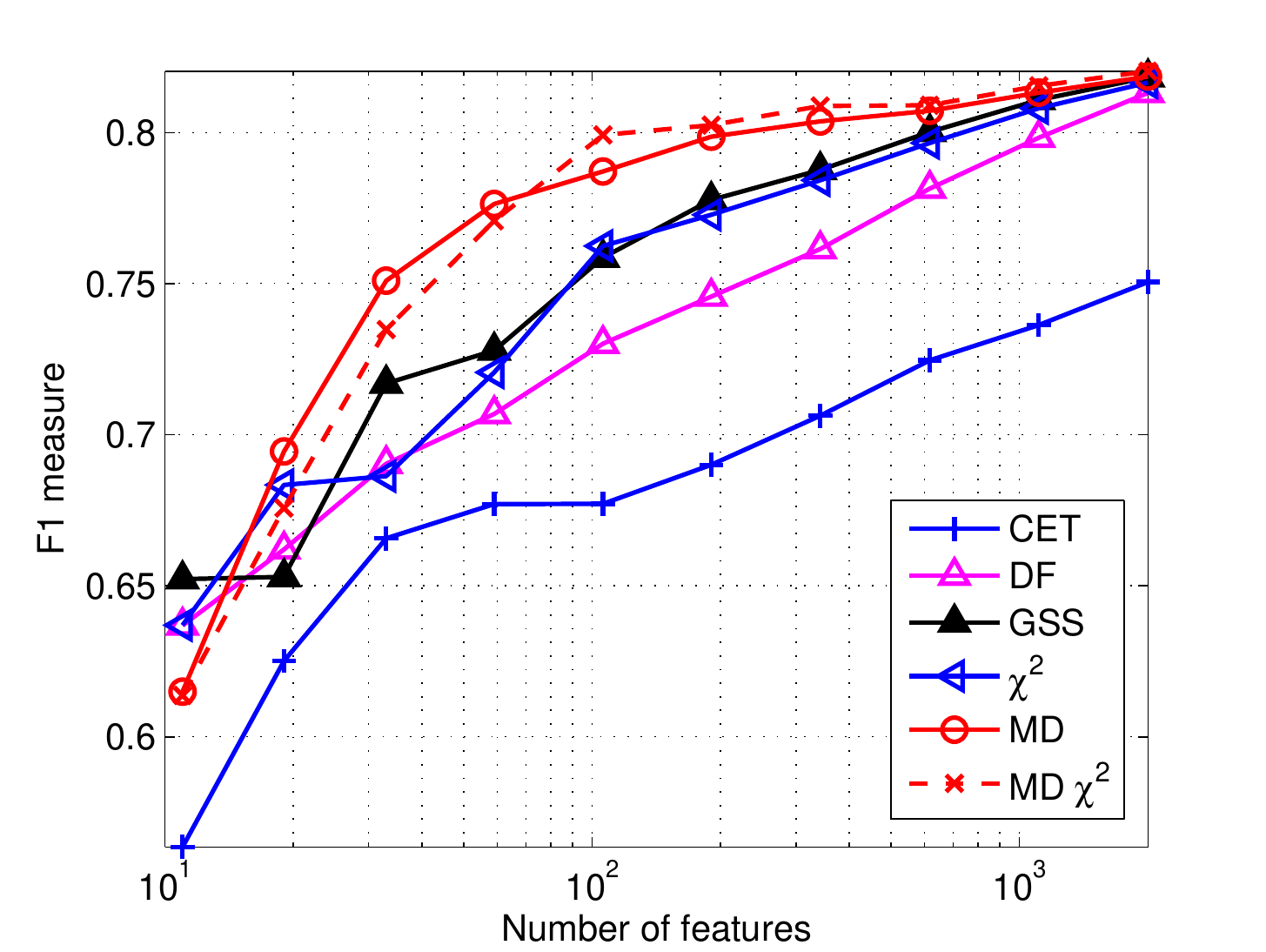}}%
  
  \caption{The results of accuracy and F1 measure on the data sets of (a). \textsc{Reuters-10}, (b). \textsc{Reuters-20}, and (c). \textsc{Reuters-30}, when naive Bayes is used as the classifier.}
  \label{Reuters-all}
\end{figure*}

We first test these feature selection approaches when naive Bayes is used as the classifier. Fig. \ref{20news_all} shows the results on the \textsc{20-Newsgroups} data set. It can be shown that the performance is improved when more features are selected. The proposed two approaches commonly perform better than others. As one can see in the zoomed-in figure, the proposed \textit{MD} method performs better than the others. The \textit{DF} method is the worst one for this data set. 
We further show the results in Fig. \ref{alt_comp} on the data set of \textsc{alt-comp} that is a subset of \textsc{20-Newsgroups} with the categories \textit{alt.*} and the categories \textit{comp.*}. For this data set, the \textit{MD-$\chi^2$} is the best one among all others. The comparison of the F1 measure on \textsc{alt-comp} is given in Fig. \ref{alt_comp}(b) and shows that the proposed \textit{MD} and its asymptotic $\chi^2$ statistic are the best two approaches. Comparing the results in \ref{20news_all} and Fig. \ref{alt_comp}, there is a significant difference on the performance behavior, although the \textsc{alt-comp} is a subset of \textsc{20-Newsgroups}. For example, the accuracy of the \textsc{20-Newsgroups} data set is even lower than $20\%$ when a small subset of features are selected, while the accuracy of the \textsc{alt-comp} data set is higher than $80\%$.  This might indicate a diverse feature characteristic of the \textsc{20-Newsgroups} data set. In the \textsc{20-Newsgroups} data set, some topics belong to the same category and are very closely related to each other, e.g., rec.sport.baseball and rec.sport.hockey, comp.sys.mac.hardware and comp.sys.ibm.hardware, etc. We also notice that the \textit{MD-$\chi^2$} method performs better than \textit{MD} method in Fig. \ref{alt_comp}, which might not be always true for other data sets since the \textit{MD-$\chi^2$} method is based on the asymptotic distribution of the statistic used in \textit{MD} method. One possible explanation is that the correlation among some words holds more discriminative information, while the \textit{MD} method assumes words are independent to each other. However, the asymptotic distribution may still hold because of the law of large number. The theoretical supports to determine which method performs better need further study.

Fig. \ref{Reuters-all} shows the comparison results on three data sets in \textsc{Reuters}: \textsc{Reuters-10}, \textsc{Reuters-20}, and \textsc{Reuters-30}. It can be shown that our proposed two approaches with the first 200 selected features can achieve the similar performance as other four  existing approaches with the first 1000 selected features. Moreover, as seen in Fig. \ref{Reuters-all}(a), (b) and (c), the performance improvement of the proposed two approaches is increased in comparison with other methods, when more categories are considered.

\begin{figure}[htp]
\captionsetup[subfigure]{labelformat=empty}
  \centering
  \subcaptionbox{(a). Accuracy for \textsc{TDT2-10} \label{TDT2:a}}{\includegraphics[width=2.4in]{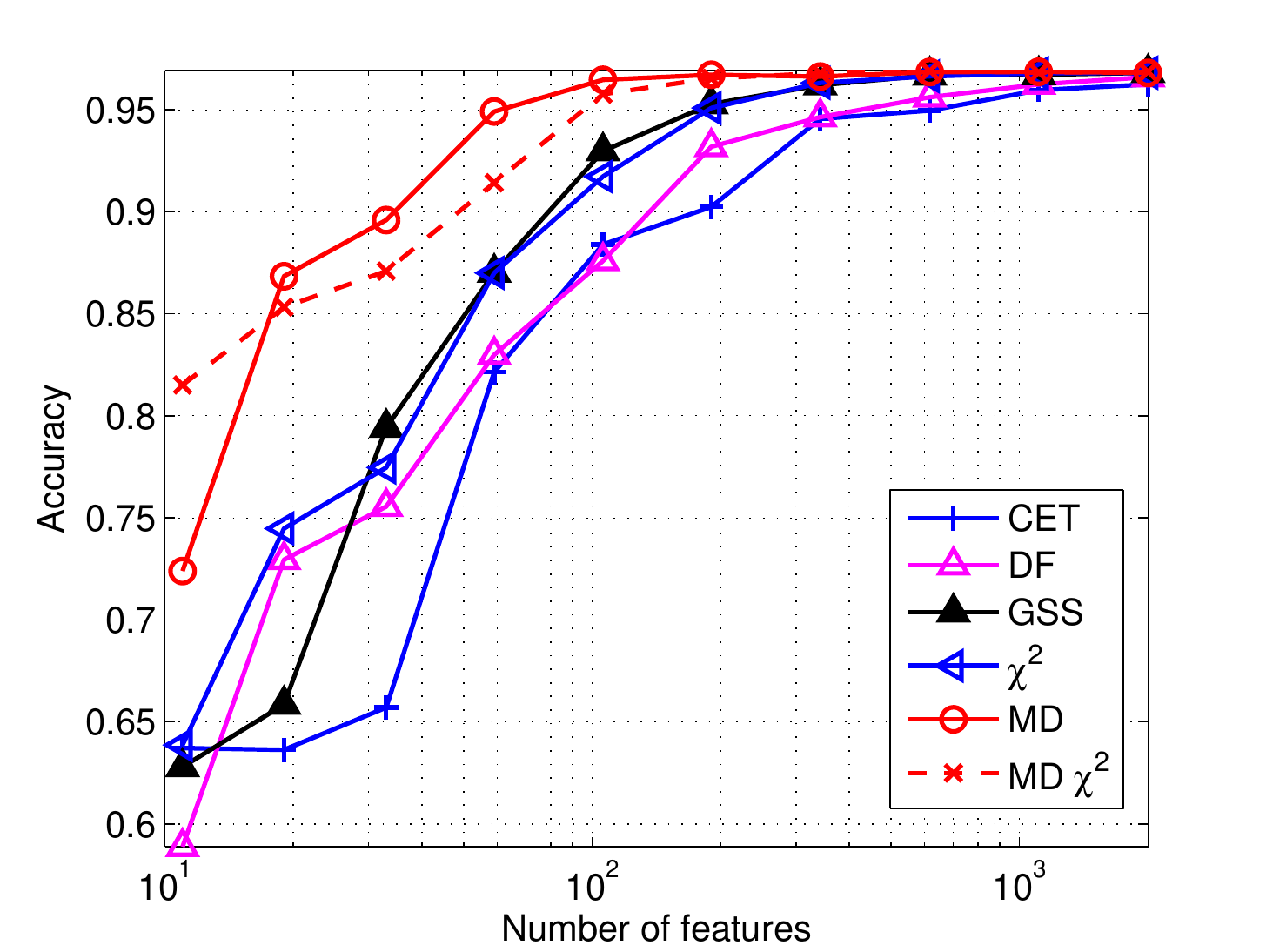}}
  
  \subcaptionbox{(b). F1 measure for \textsc{TDT2-10} \label{TDT2:b}}{\includegraphics[width=2.4in]{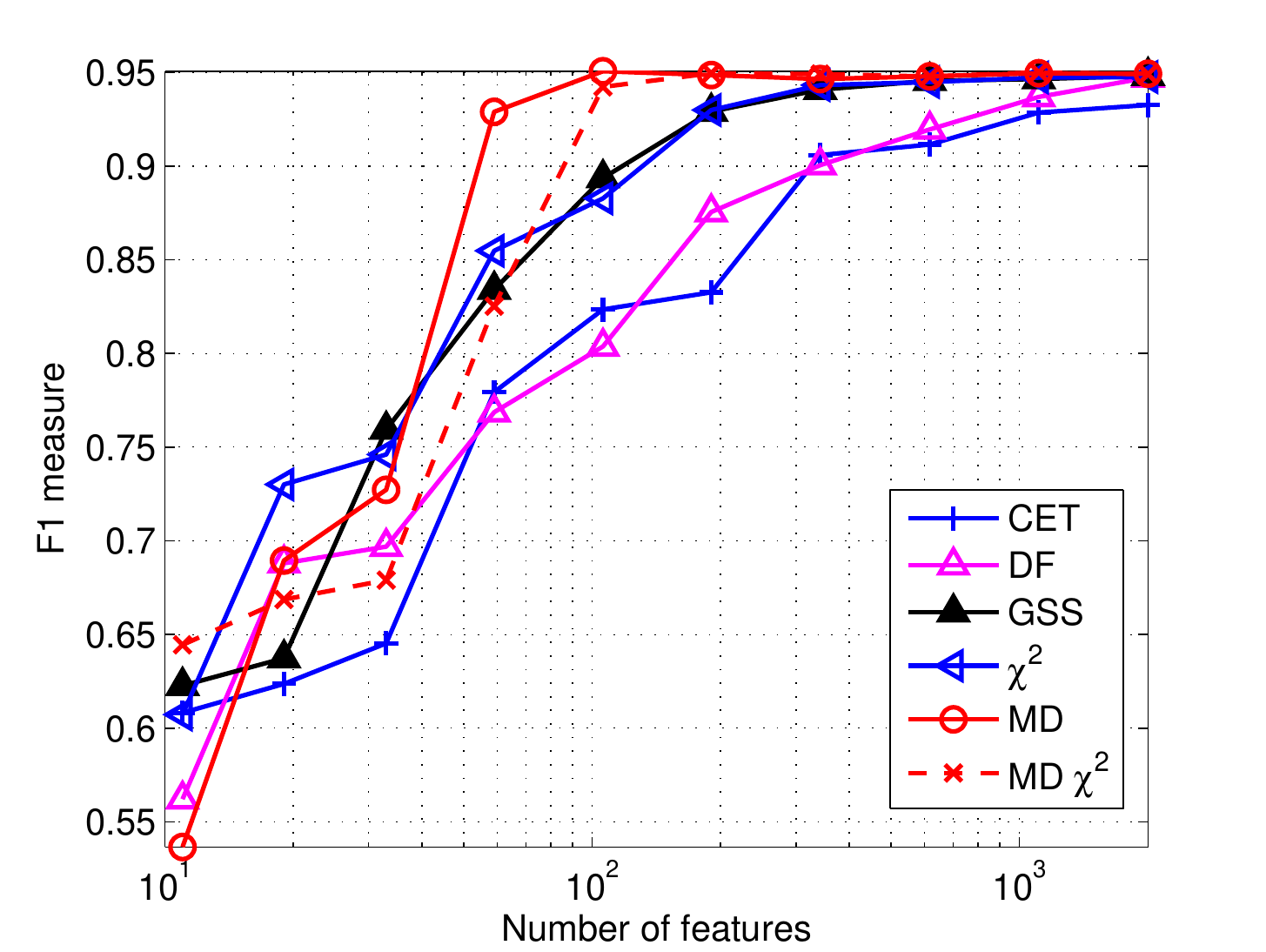}}\hspace{0.1em}%
  
  \caption{Performance comparisons of feature selection methods on the \textsc{TDT2} data set: (a) accuracy, (b) F1 measure, when naive Bayes is used as the classifier.}
  \label{TDT2}
\end{figure}

Fig. \ref{TDT2} shows results for the \textsc{TDT2} data set. As the classification tasks on this data set are performed with the scheme of 10-fold cross validation, all the results in Fig. \ref{TDT2} are averaged across 10 runs. As shown in Fig. \ref{TDT2}(a), the proposed \textit{MD} and \textit{MD-$\chi^2$} outperform all the others with respect to the metric of accuracy. It is interesting to notice that, when the first $100$ features are selected, the \textit{MD} obtains the accuracy of $96.46\%$, and the \textit{MD-$\chi^2$} has $95.76\%$. For other methods, such as \textit{GSS} and \textit{$\chi^2$}, the first $1000$ features need to be selected to achieve the same classification accuracy. 

\begin{figure*}[htp]
\captionsetup[subfigure]{labelformat=empty}
  \centering
  \subcaptionbox{(a-1). Accuracy for \textsc{Reuters-10} \label{Reuters_svm:a1}}{\includegraphics[width=2.3in]{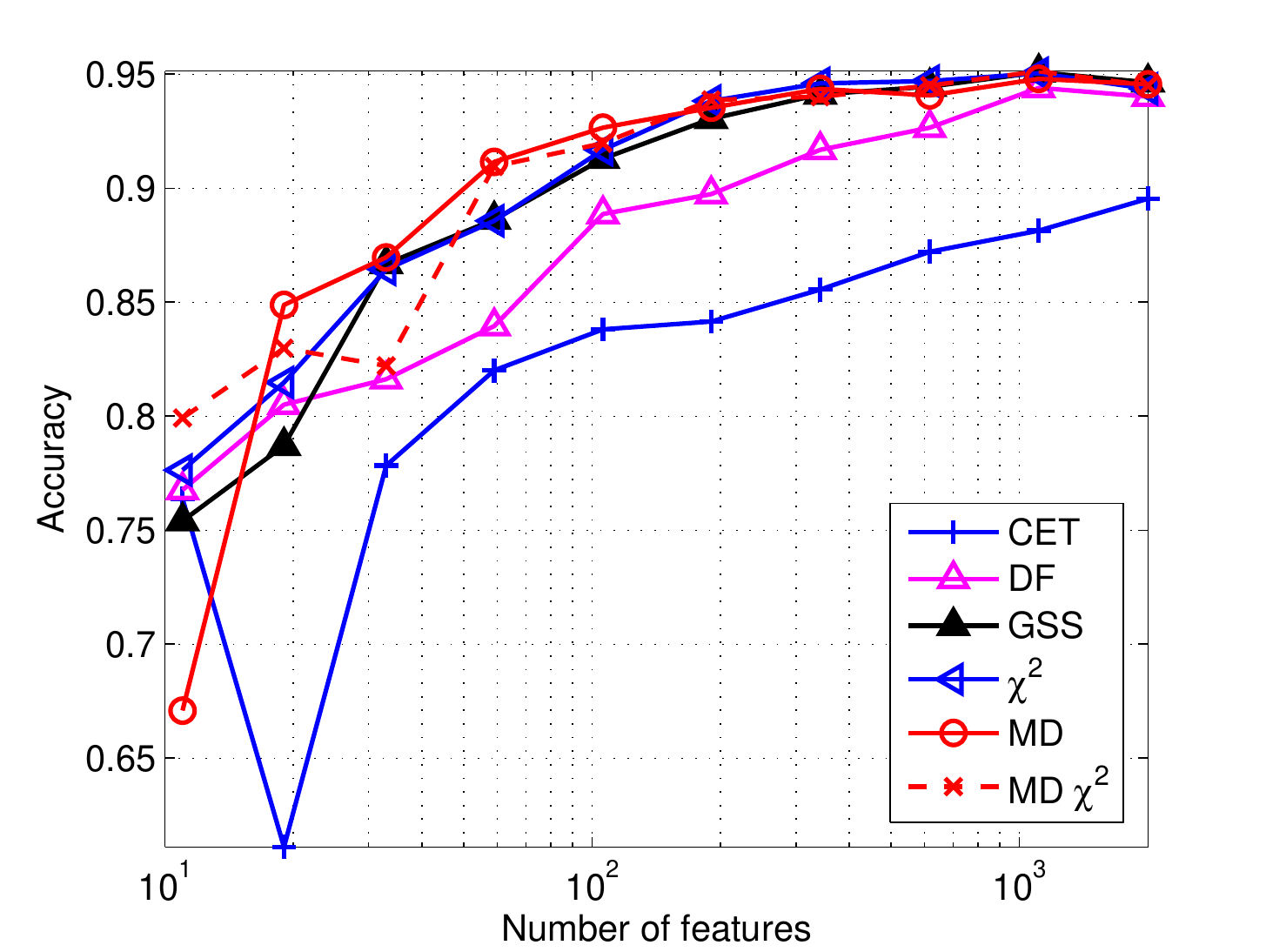}}\hspace{0.1mm}%
  \subcaptionbox{(a-2). F1 measure for \textsc{Reuters-10} \label{Reuters_svm:a2}}{\includegraphics[width=2.3in]{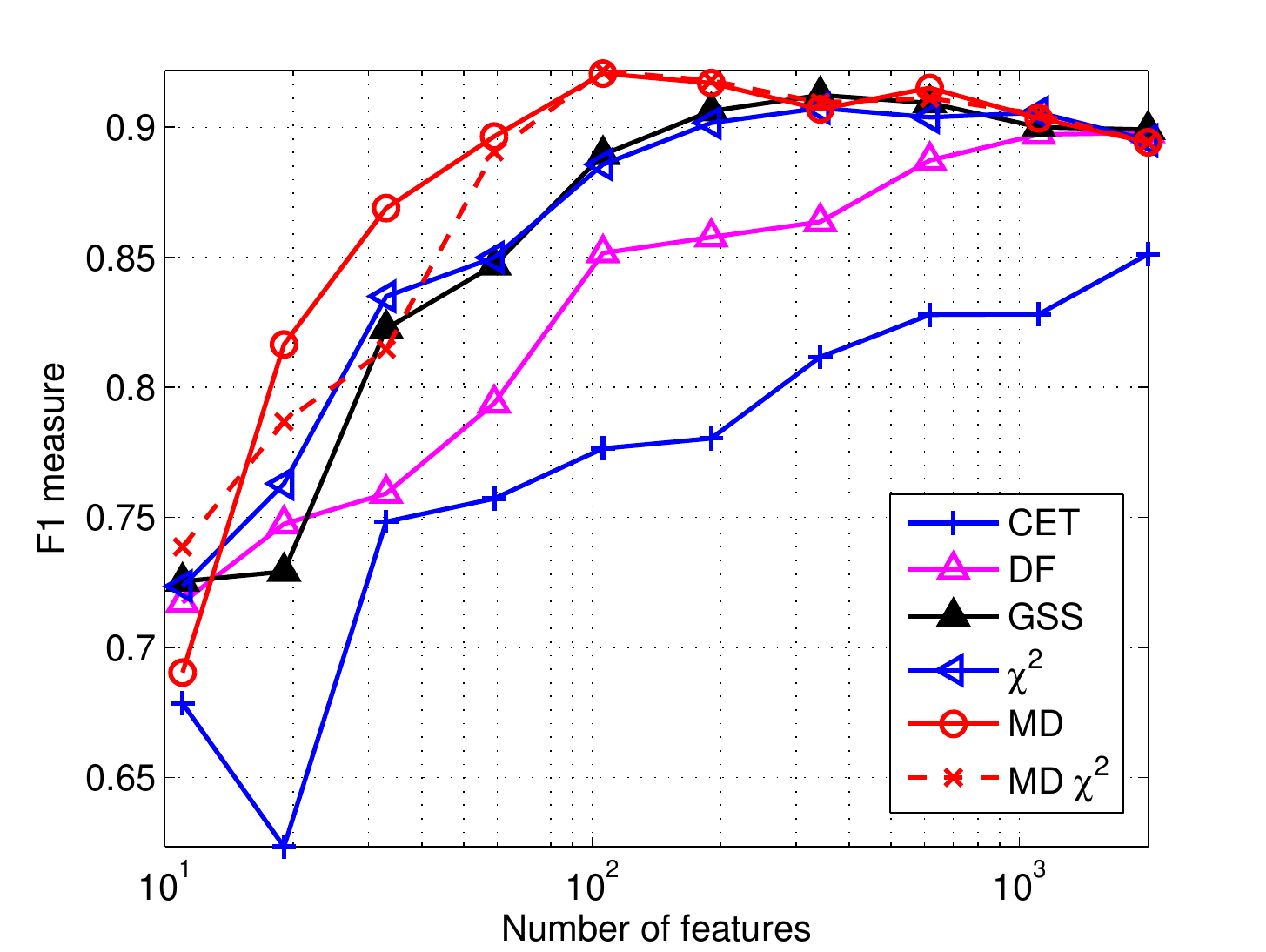}}\hspace{0.1em}%
  
  \subcaptionbox{(b-1). Accuracy for \textsc{Reuters-20} \label{Reuters_svm:b1}}{\includegraphics[width=2.3in]{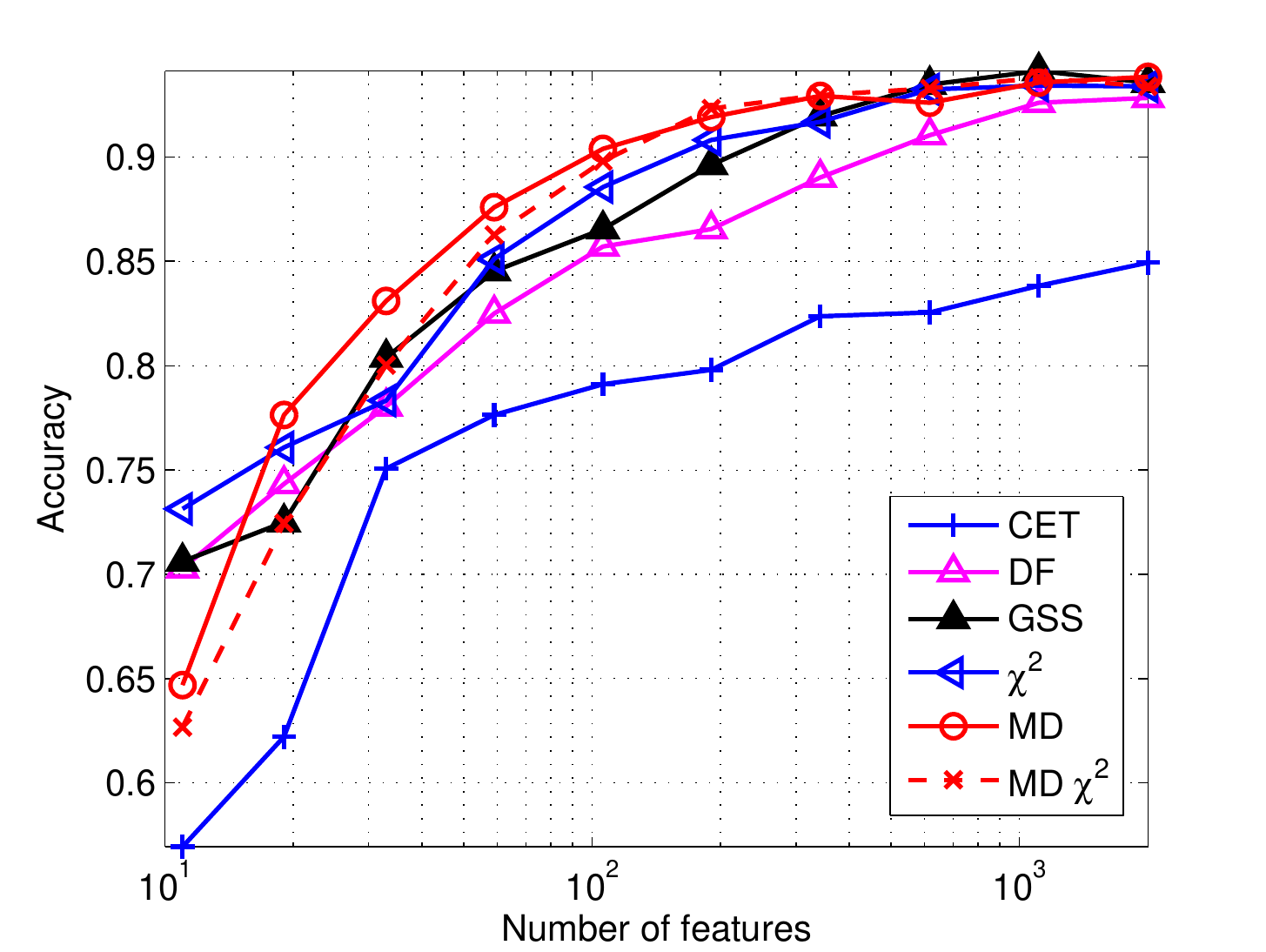}}\hspace{0.1em}%
  \subcaptionbox{(b-2). F1 measure for \textsc{Reuters-20}  \label{Reuters_svm:b2}}{\includegraphics[width=2.3in]{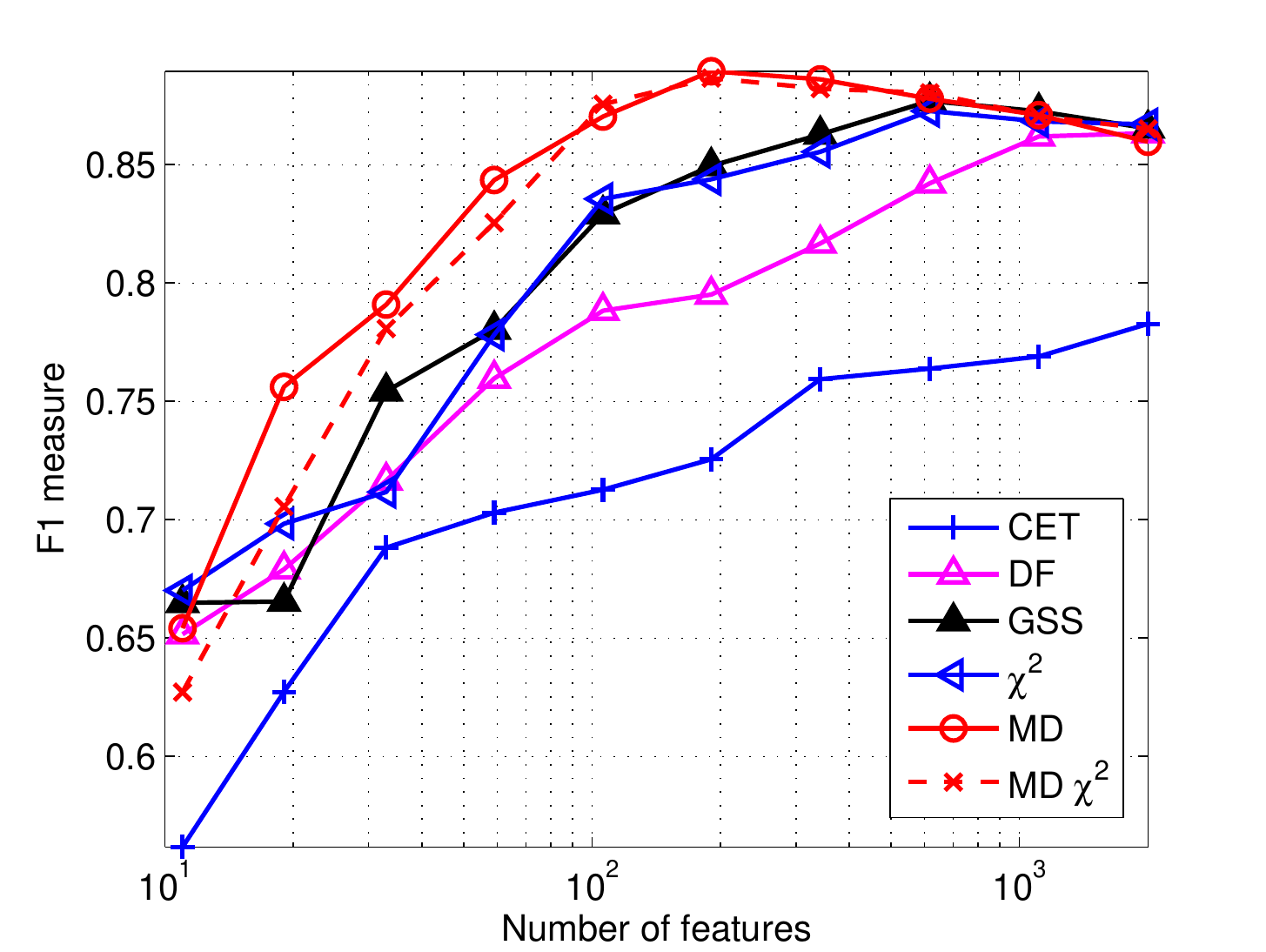}}
  
  \subcaptionbox{(c-1). Accuracy for \textsc{Reuters-30} \label{Reuters_svm:c1}}{\includegraphics[width=2.3in]{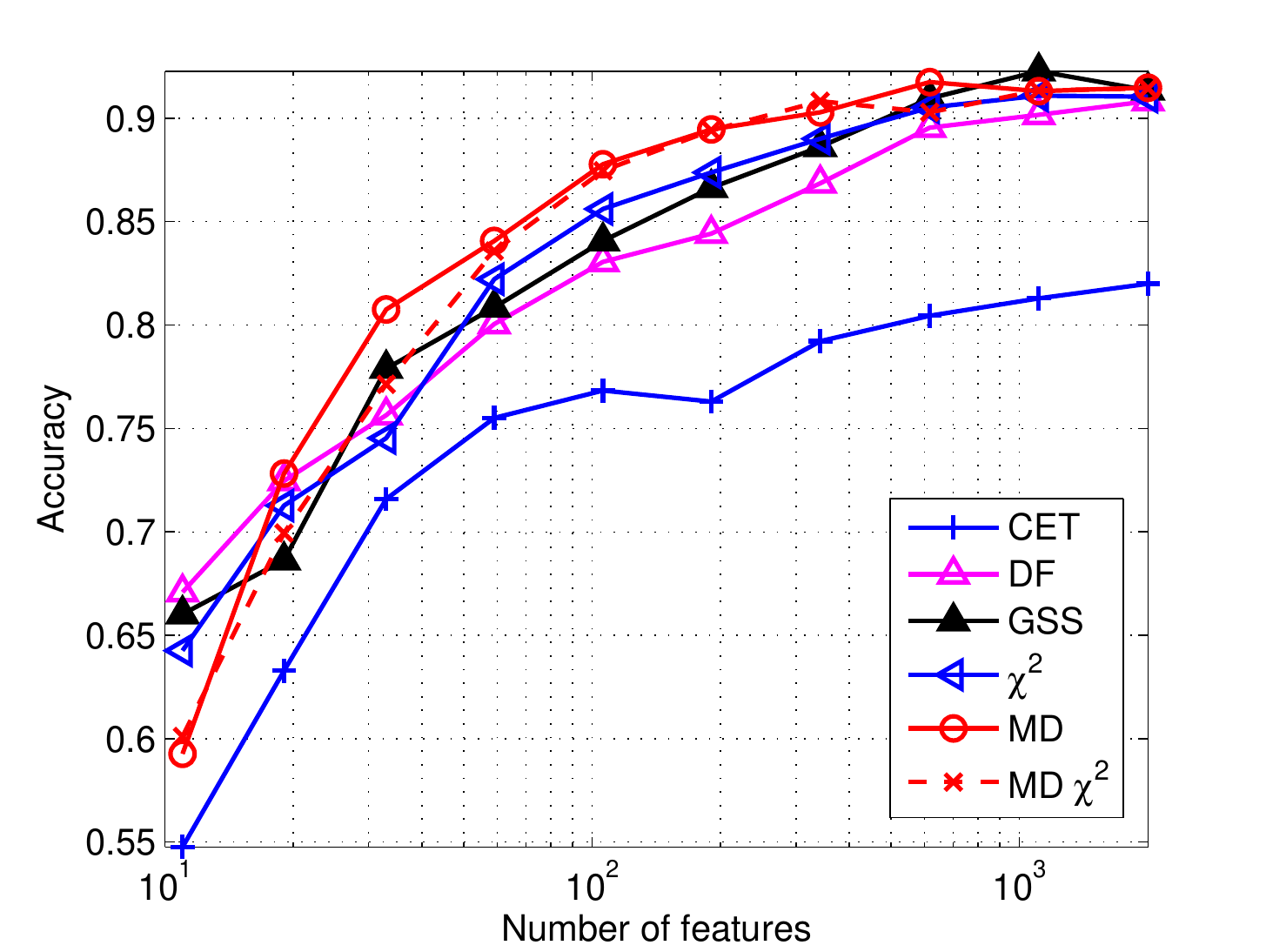}}\hspace{0.1em}%
  \subcaptionbox{(c-3). F1 measure for \textsc{Reuters-30} \label{Reuters_svm:c2}}{\includegraphics[width=2.3in]{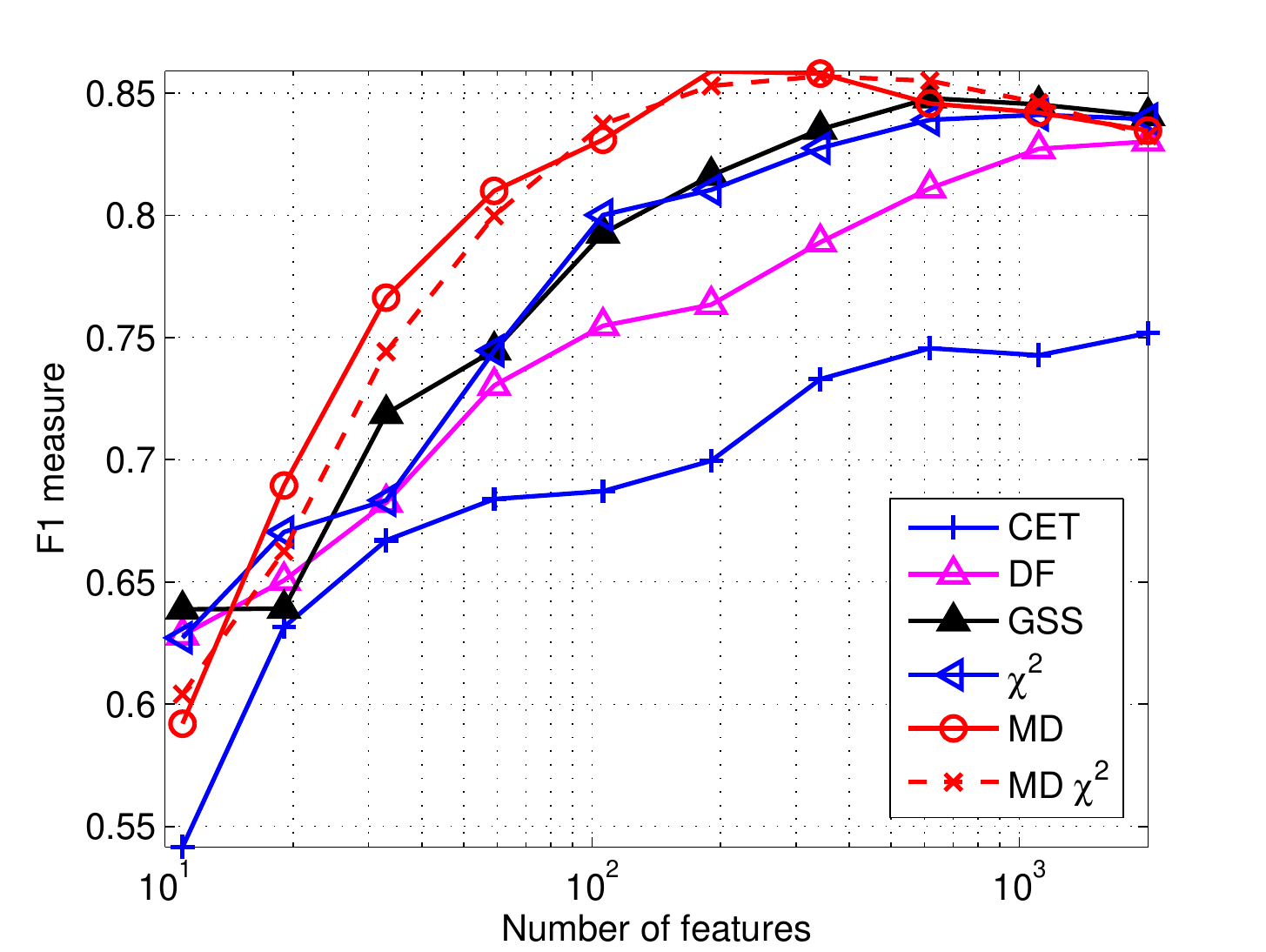}}%
  
  \caption{The results of accuracy and F1 measure on the data sets of (a). \textsc{Reuters-10}, (b). \textsc{Reuters-20}, and (c). \textsc{Reuters-30}, when SVM is used as the classifier.}
  \label{Reuters-svm}
\end{figure*}

We also test our proposed feature selection approaches compared with the previously existing feature selection methods, when SVM is used as classifier for text categorization. Fig. \ref{Reuters-svm} shows the classification results on three \textsc{Reuters} data sets for SVM, in which the performance improvement of our proposed two approaches can be also seen. It shows that our proposed two approaches perform at least as well as previously existing methods at a very small feature size, and are consistently better when the feature size increases.  

\section{Conclusions and Future Works}
We have introduced new feature selection approaches based on the information measures for naive Bayes classifiers, aiming to select the features that offer the maximum discriminative capacity for text classification. We have also derived the asymptotic distributions of these measures, which leads to the other version of the Chi-square statistic approach for feature selection. Compared with the existing feature selection approaches that rank the features by only exploring the intrinsic characteristics of data without considering the learning algorithm for classification, our proposed approaches involve the learning model in the feature filtering process, which provides us a theoretical way to analyze the optimality of the selected features. The experiments we have conducted on several benchmarks have demonstrated their promising performance improvement compared with the previously existing feature selection approaches.

For future work, we will analyze the feature dependence and develop feature selection algorithms by weighting each individual features\cite{xue2009distributional}\cite{kim2006some}, aiming to maximize the discriminative capacity. Furthermore, we will incorporate our feature selection approaches into other advanced machine learning algorithms such as imbalanced learning \cite{tang2015kernel}\cite{he2009learning} and partial learning model \cite{tang2014hybrid}\cite{tang2015parametric} to enhance the learning for rare categories.

\section*{Acknowledgment}
This research was partially supported by National Science Foundation (NSF) under grant ECCS 1053717 and CCF 1439011, and the Army Research Office under grant W911NF-12-1-0378.

\bibliographystyle{IEEEtran}

\end{document}